\documentclass[journal]{IEEEtran}
\usepackage{csquotes}
\usepackage{graphicx}
\usepackage{soul}
\usepackage{float}
\usepackage{multirow}
\DeclareUnicodeCharacter{0301}{'{e}}
\usepackage{amsmath}
\usepackage{tikz}
\usepackage{graphicx}
\usepackage{subcaption}
\usetikzlibrary{shapes.geometric, arrows.meta, positioning, backgrounds, fit,calc}
\usepackage[normalem]{ulem}
\usepackage{xspace}
\newcommand{\latinphrase}[1]{\textit{#1}}
\newcommand{\etal}{\latinphrase{et~al.}\xspace}
\newcommand{\ie}{\latinphrase{i.e.}\xspace}

\newcommand{\eg}{\latinphrase{e.g.}\xspace}

\usepackage[pagebackref=false,breaklinks=true,colorlinks,bookmarks=false]{hyperref}
\makeatother
\begin{document}

\title{Towards Autonomous Riding: A Review of Perception, Planning, and Control in Intelligent Two-Wheelers}

\author{
Mohammed Hassanin, Mohammad Abu Alsheikh, Carlos C. N. Kuhn, Damith Herath, Dinh Thai Hoang, and Ibrahim Radwan
\thanks{M. Hassanin, M. A. Alsheikh, C. C. N. Kuhn, D. Herath and I. Radwan are with the Faculty of Science \& Technology, University of Canberra, Canberra, ACT 2617, Australia, emails: {Mohammed.Hassanin; Mohammad.Abualsheikh;  Damith.Herath; Ibrahim.Radwan}@canberra.edu.au.}
\thanks{C. C. N. Kuhn is with OpenSI - Faculty of Science and Technology, University of Canberra, Canberra, ACT 2617, Australia, email: carlos.noschangkuhn@canberra.edu.au.}
\thanks{D. T. Hoang is with the School of Electrical and Data Engineering, University of Technology Sydney, NSW 2007, Australia, email: hoang.dinh@uts.edu.au.}
}
\maketitle
\begin{abstract}

The rapid adoption of micromobility solutions, particularly two-wheeled vehicles like e-scooters and e-bikes, has created an urgent need for reliable autonomous riding (AR) technologies. While autonomous driving (AD) systems have matured significantly, AR presents unique challenges due to the inherent instability of two-wheeled platforms, limited size, limited power, and unpredictable environments, which pose very serious concerns about road users' safety. This review provides a comprehensive analysis of AR systems by systematically examining their core components, perception, planning, and control, through the lens of AD technologies. We identify critical gaps in current AR research, including a lack of comprehensive perception systems for various AR tasks, limited industry and government support for such developments, and insufficient attention from the research community. The review analyses the gaps of AR from the perspective of AD to highlight promising research directions, such as multimodal sensor techniques for lightweight platforms and edge deep learning architectures. By synthesising insights from AD research with the specific requirements of AR, this review aims to accelerate the development of safe, efficient, and scalable autonomous riding systems for future urban mobility.

\end{abstract}

\IEEEpeerreviewmaketitle

\section{Introduction}

\begin{figure}
\centering
\scalebox{0.6}{
\begin{tikzpicture}[
  every node/.style={font=\small, align=center},
  challenge/.style={circle, draw, minimum size=3.2cm, inner sep=0, fill=white}
]

\def\coneColors{{"red!20","orange!20","yellow!20","green!20","blue!20","purple!20","pink!20","teal!20"}}
\newcounter{colorindex}\setcounter{colorindex}{0}

\coordinate (poor-c)    at (-3.6,  5.5);
\coordinate (traffic-c) at ( 0,     6  );
\coordinate (drivers-c) at ( 4,     5  );
\coordinate (potholes-c)at (-5,     1  );
\coordinate (balance-c) at ( 5,     0.3);
\coordinate (blind-c)   at (-4.5,  -2.8);
\coordinate (ped-c)     at ( 3.7,  -3.5);
\coordinate (battery-c) at ( 0,    -5  );

\foreach \pos in {poor-c,traffic-c,drivers-c,potholes-c,balance-c,blind-c,ped-c,battery-c} {
  \pgfmathparse{\coneColors[\thecolorindex]}
  \edef\currentcolor{\pgfmathresult}
  \path let
    \p1 = ($(\pos)-(0,0)$),
    \n1 = {atan2(\y1,\x1)}
  in
    coordinate (\pos-left)  at ($(\pos)+(\n1+90:1.6cm)$)
    coordinate (\pos-right) at ($(\pos)+(\n1-90:1.6cm)$);
  \fill[\currentcolor,opacity=0.4]
    (0,0) -- (\pos-left) -- (\pos-right) -- cycle;
  \stepcounter{colorindex}
}

\node[challenge] (poor)    at (poor-c)    {\includegraphics[width=2.2cm]{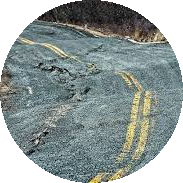}};
\node[below=0.15cm of poor]    {\textbf{Poor Infrastructure}};

\node[challenge] (traffic) at (traffic-c) {\includegraphics[width=2.2cm]{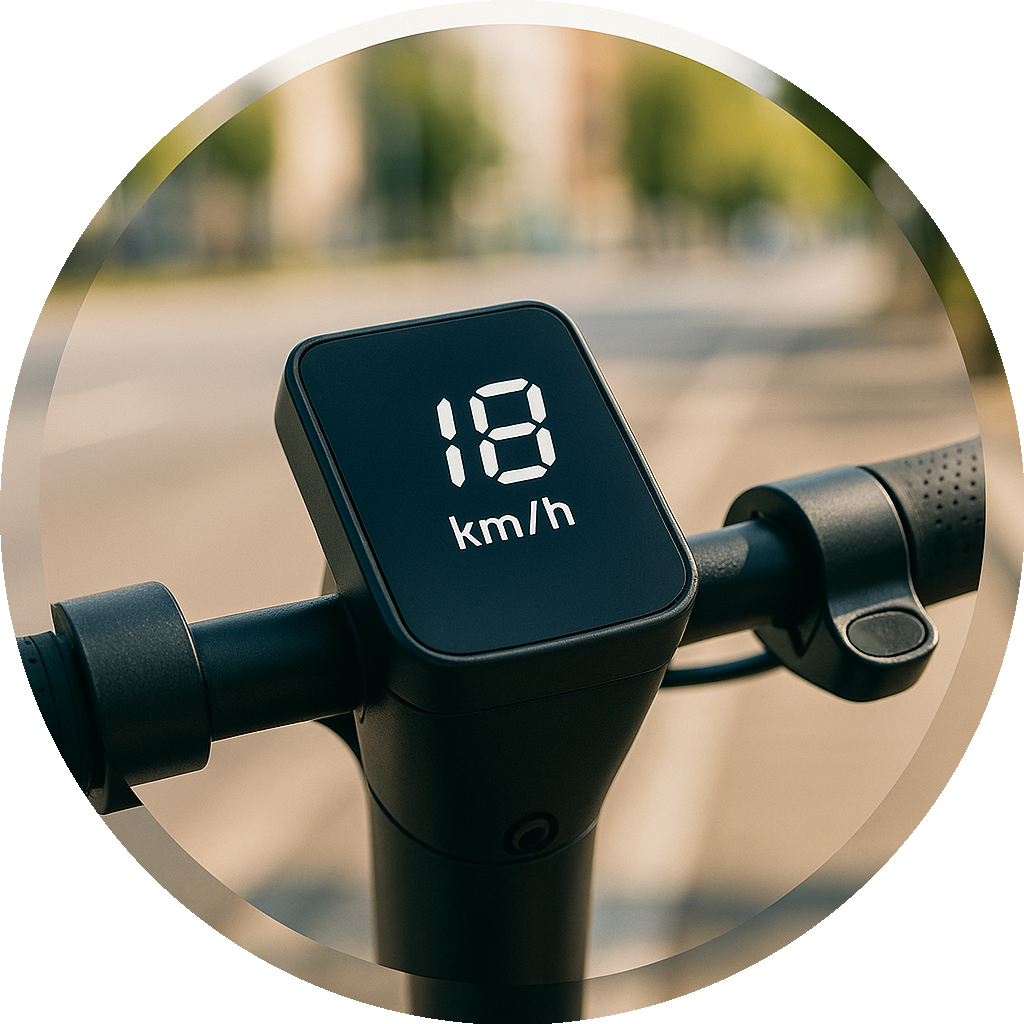}};
\node[below=0.15cm of traffic] {\textbf{Speed Control}};

\node[challenge] (drivers) at (drivers-c) {\includegraphics[width=2.2cm]{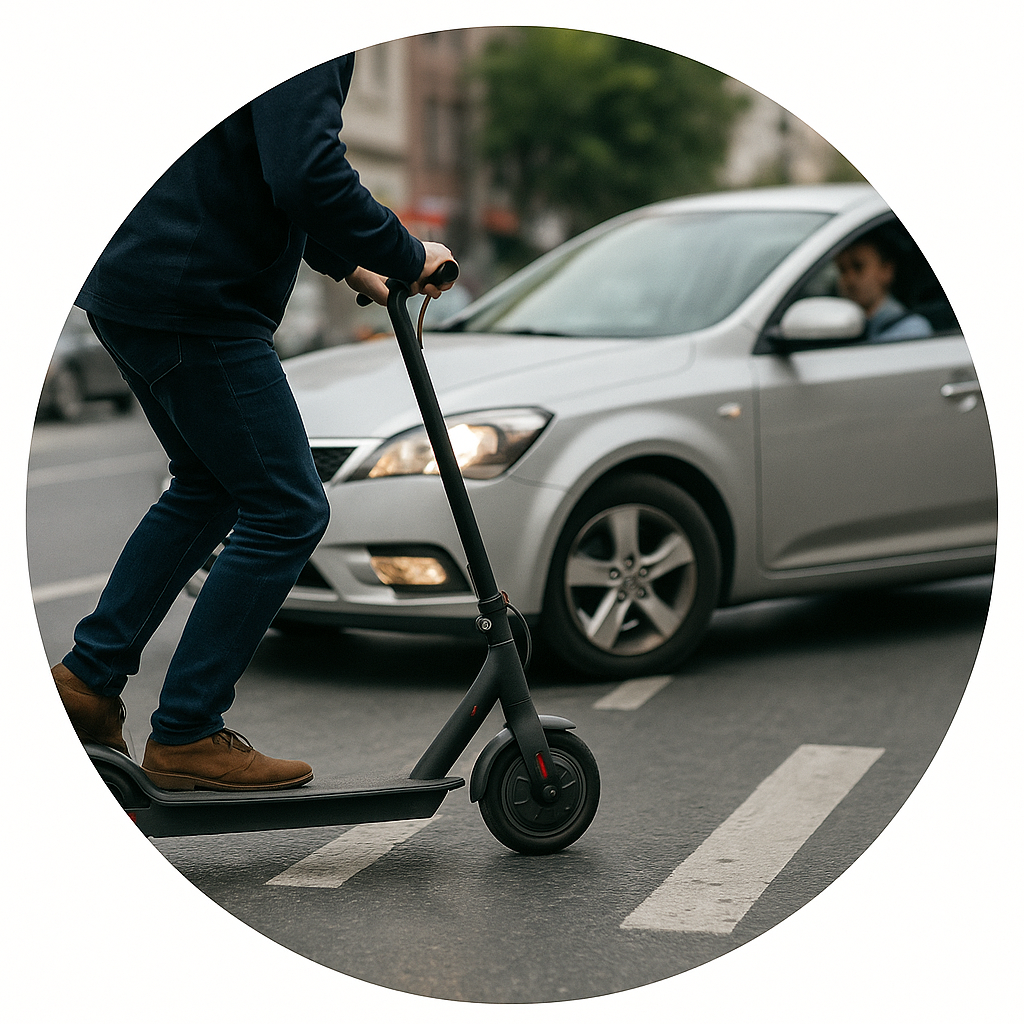}};
\node[below=0.15cm of drivers] {\textbf{Road Users' Behaviours}};

\node[challenge] (potholes) at (potholes-c) {\includegraphics[width=2.2cm]{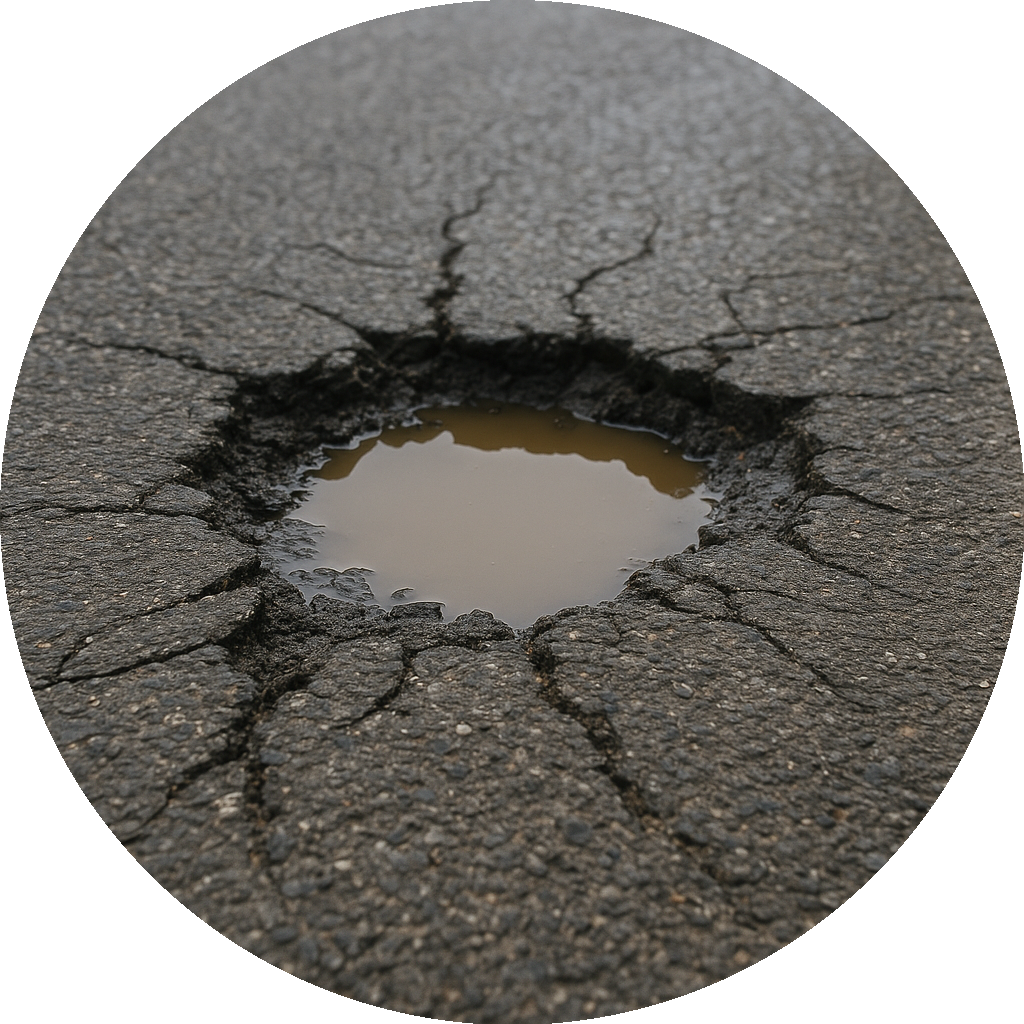}};
\node[below=0.15cm of potholes] {\textbf{Potholes}};

\node[challenge] (balance)  at (balance-c) {\includegraphics[width=2.2cm]{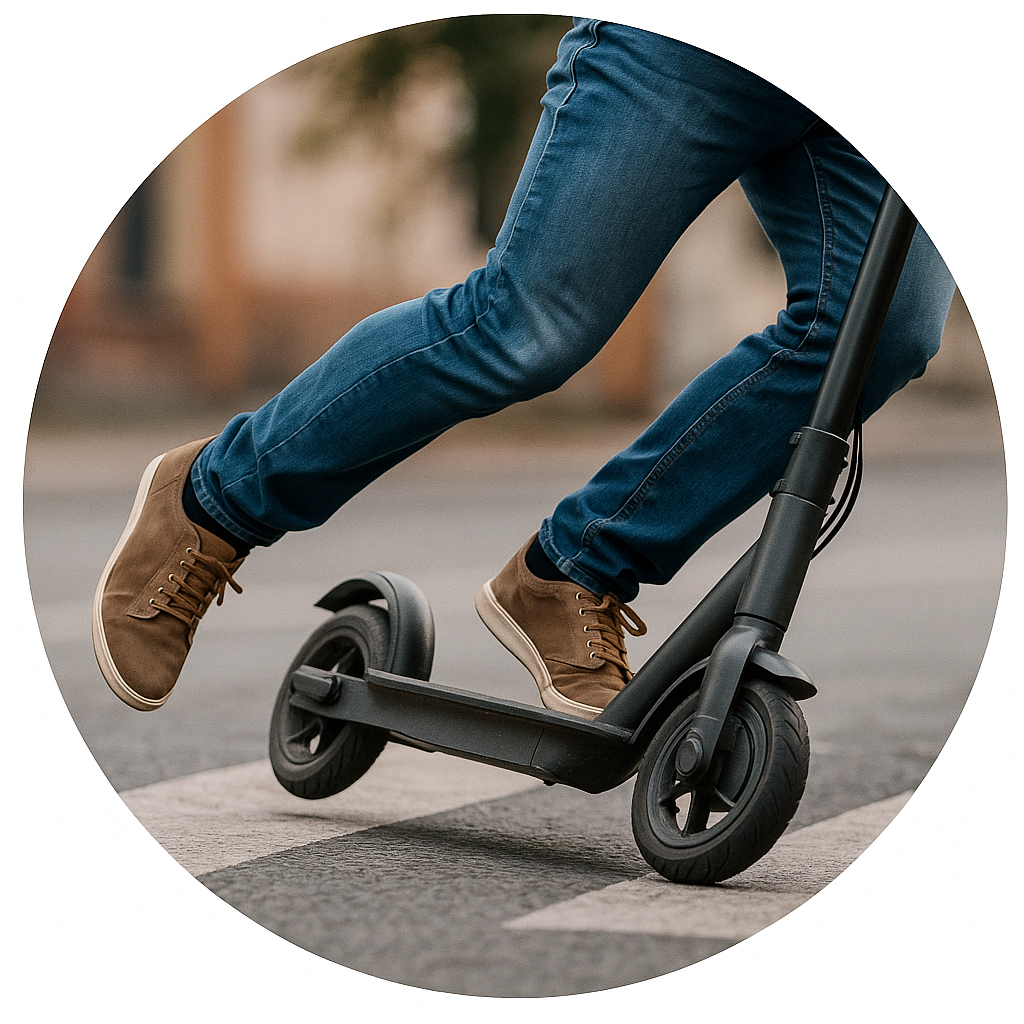}};
\node[below=0.15cm of balance]  {\textbf{Balance}};

\node[challenge] (blind)    at (blind-c)   {\includegraphics[width=2.2cm]{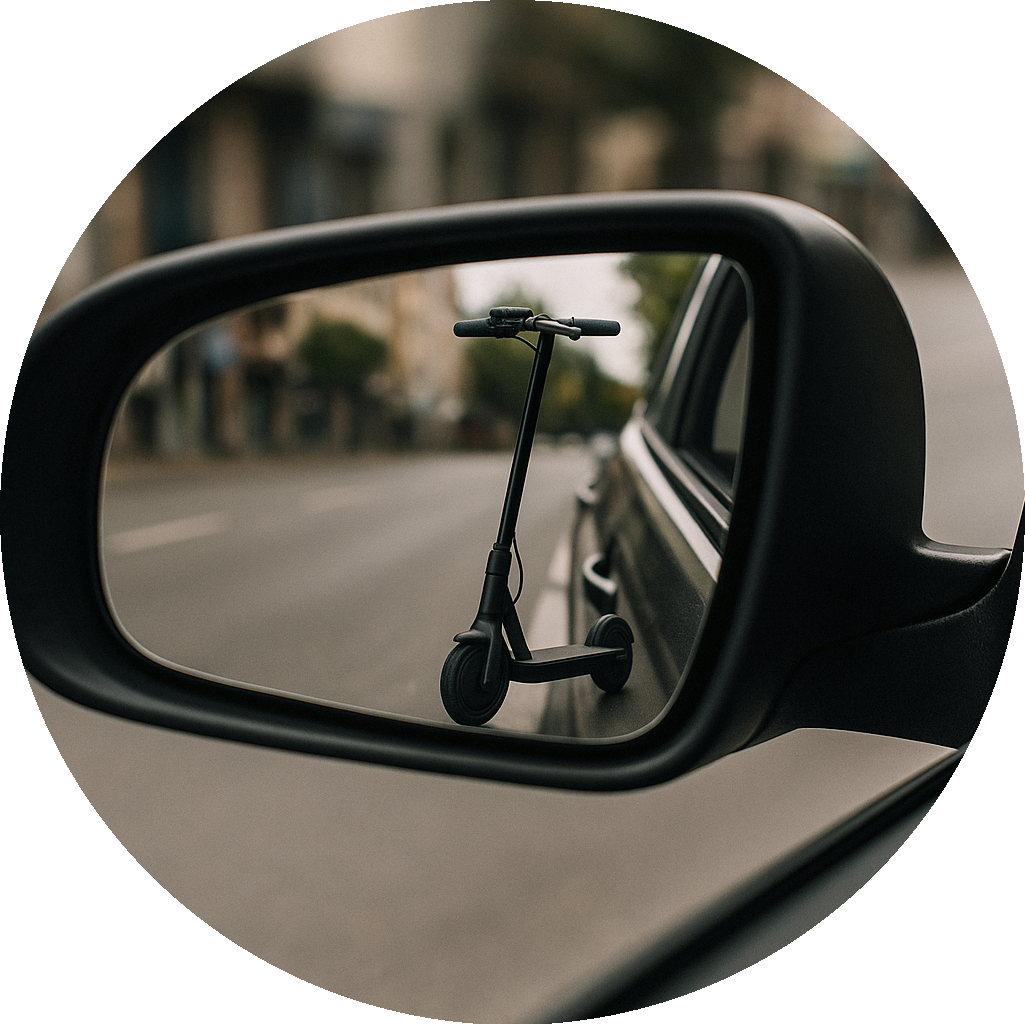}};
\node[below=0.15cm of blind]    {\textbf{Blind Spots}};

\node[challenge] (ped)      at (ped-c)     {\includegraphics[width=2.2cm]{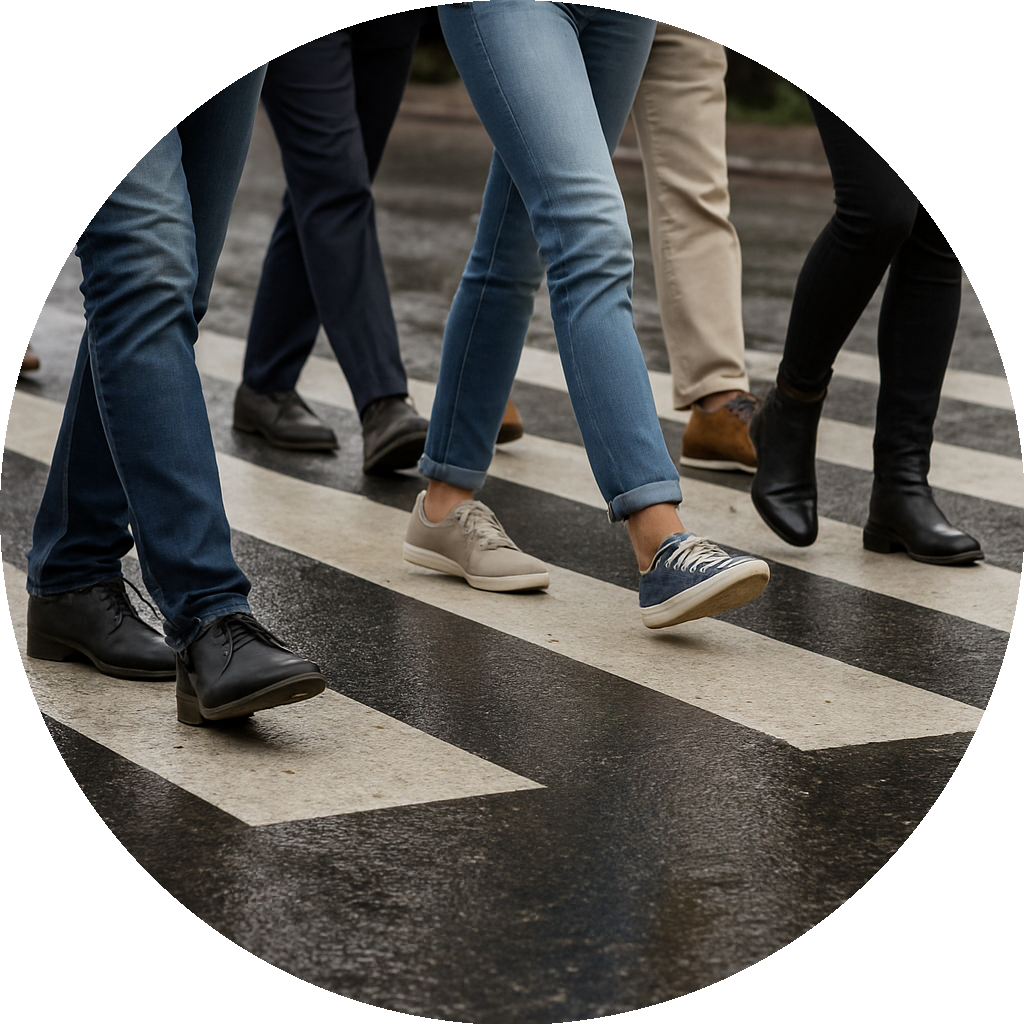}};
\node[below=0.15cm of ped]      {\textbf{Crossings}};

\node[challenge] (battery)  at (battery-c) {\includegraphics[width=2.2cm]{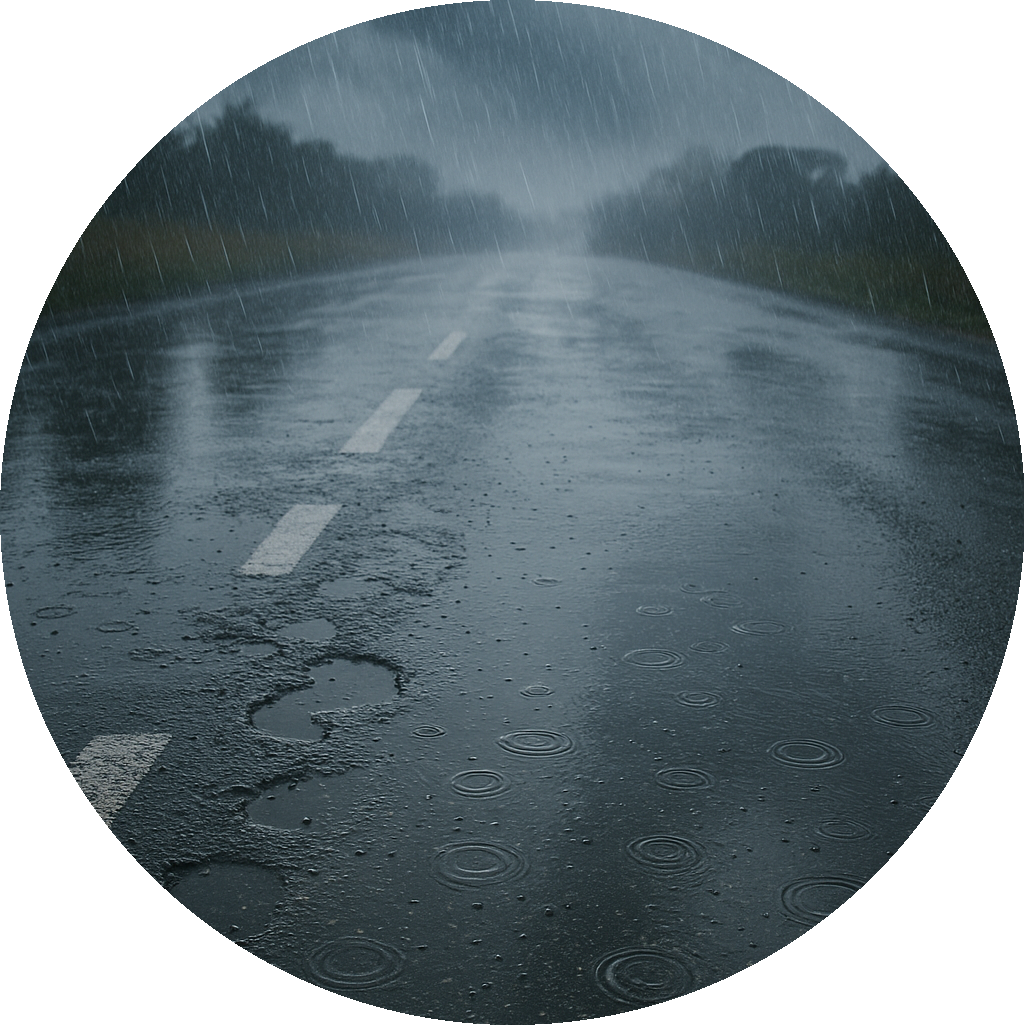}};
\node[below=0.15cm of battery]  {\textbf{Adverse Weather}};

\node (scooterimg) at (0,0)
  {\includegraphics[width=5cm]{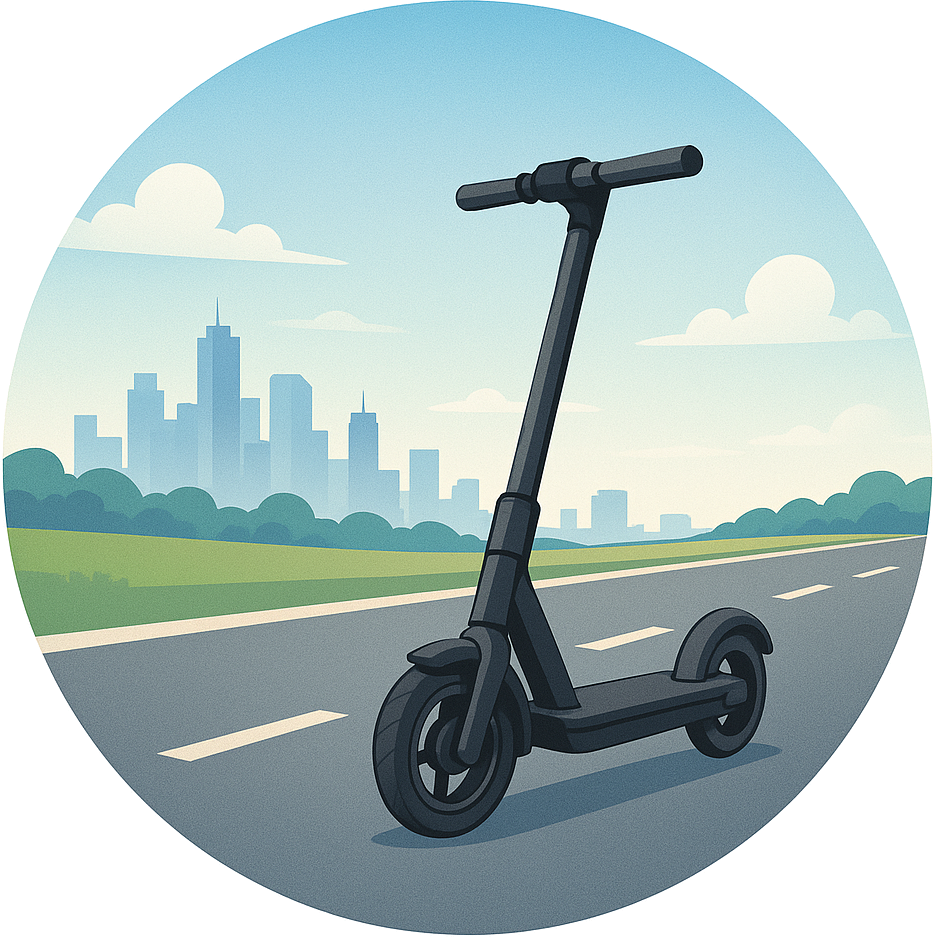}};

\end{tikzpicture}
}
\caption{Visual explanation of the key safety challenges of two-wheel vehicles. Conical sectors illustrate eight primary challenges: poor infrastructure, speed control, road users' behaviour, potholes, balance, blind spots, pedestrian crossings, and adverse weather, centred on two-wheelers. These challenges require high attention from the research community to cope with the expected risks of two-wheelers.}
\label{fig:challenges}
\end{figure}
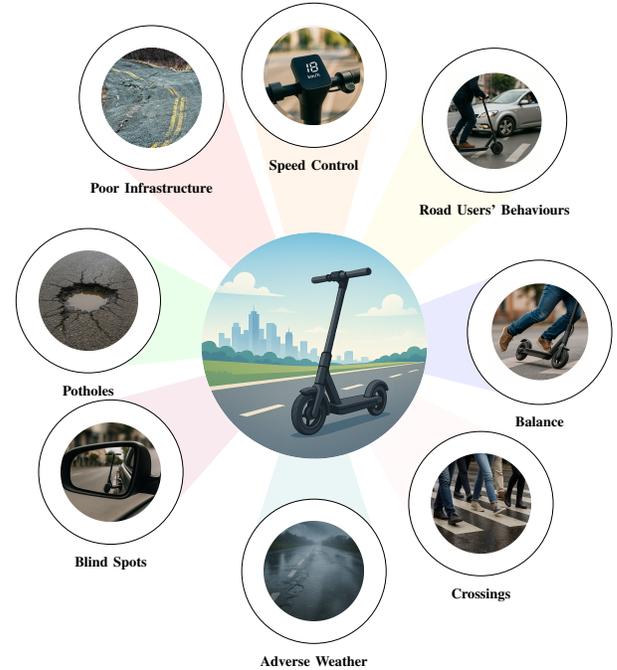

\IEEEPARstart{M}icromobility riding has recently been witnessing an increase due to their potential to improve urban transport efficiency, reduce emissions, and enhance road safety \cite{imove_micromobility}. Recent studies in this area play a critical role in bridging the gap between autonomous driving and micromobility riding. However, most of the existing works focused on narrow technical topics, which do not reflect the natural necessity of such an emerging trend, nor do they draw sufficient attention from academia and industry equal to their widespread impact. The goal of this work is to provide a comprehensive overview of the state-of-the-art of AR and outline promising directions for future research in the shadow of autonomous driving. This overview compiles key ideas in literature, introduces research themes and ethical considerations, and suggests future research pathways about core technologies such as sensing, safety systems, real-time edge intelligence, and behaviour modelling. This work is expected to serve as a foundational guide linking past innovations to future advancements in micromobility safety and intelligence, keeping autonomous driving (AD) in view as a guide and as an integral part to achieve smart cities' goals.
\begin{figure*}
\centering
\scalebox{0.8}{
\includegraphics[width=\linewidth]{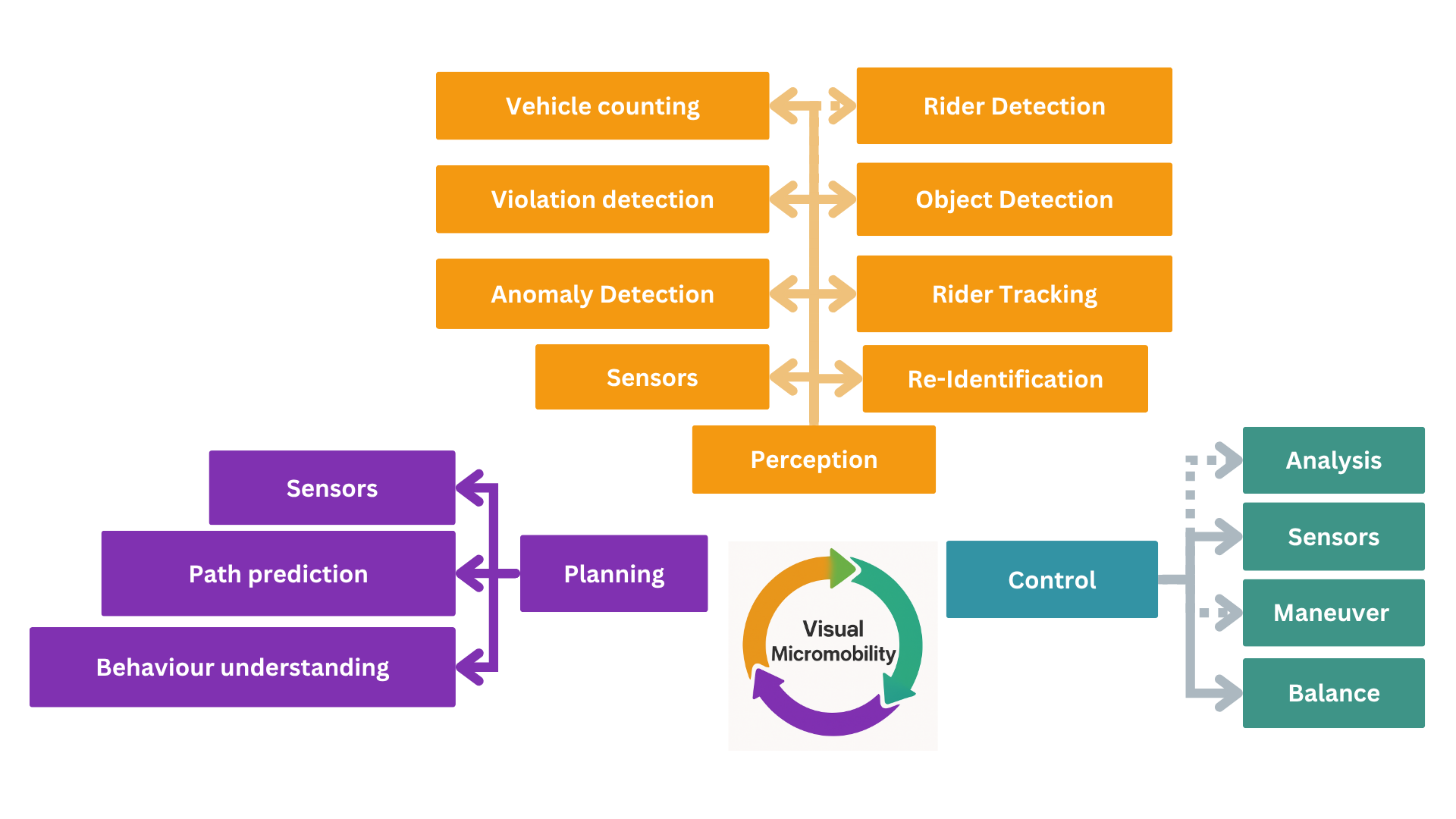}
}
\vspace{-2em}
\caption{The taxonomy of this survey, divided into three principal modules, Perception, Planning, and Control, and their key tasks. In Perception, sensor fusion, object detection, rider detection, rider tracking, Re-identification, anomaly detection, violation detection, and vehicle counting. Planning encompasses rider activity understanding, path prediction, and sensors. Control addresses manoeuvre and balance stabilisation. Solid arrows indicate areas that have been covered in the existing literature; dashed arrows highlight modules and subtasks, such as deeper behaviour analysis, and manoeuvres that remain completely unexplored.}
\label{fig:taxs}
\end{figure*}

Unlike vehicles, micromobility riders exhibit unique behaviours, such as frequent usage of pavements, inconsistent adherence to dedicated lanes, and informal lane-sharing practices. These behaviours result in unpredictable trajectories that pose new challenges for traffic systems. Despite these differences, all micromobility riders, including e-scooter, bicycle, and motorcycle riders, share common vulnerabilities due to their limited protection, physical exposure to various types of dynamic road dangers and low visibility \cite{anke2023micro}. Figure \ref{fig:challenges} summarises these challenges, which require rapid action from the research community to address them, particularly in the case of two-wheelers, which spread exponentially. Looking ahead to the future of micromobility in the context of Autonomous Vehicles (AVs) and smart city infrastructure, developing systems that can accurately detect, track, and predict the movement of such agents in dynamic driving environments \cite{yaqoob2019autonomous}. Lagging behind can lead to safety risks and reduced trust in autonomous driving. Consequently, the research community should pay attention to developing perception algorithms and context-aware interaction models to ensure safety for micromobility riders within the broader intelligent transportation systems. The progression of autonomous driving and intelligent vehicles (IVs) offers crucial insights for micromobility systems, especially as local mobility shifts towards lightweight, personalised transport \cite{sanchez2024shared}. Although breakthroughs in computer vision, sensor fusion, and neural networks have impacted the advancement of AD, micromobility is undergoing a transition, from simple pedal-powered bikes to smart devices in the context of smart cities. This leads to systems of behaviour understanding and safe mechanisms. For example, the integration of helmet detection, social alert systems, or contextual risk monitoring in micromobility can be viewed as a direct evolution of the corresponding ones in IVs.

Similar to AD, these systems integrate GPS for positioning, edge computing units for onboard data processing, and sensors such as cameras, LiDAR, ultrasonic, and IMUs to enable environment sensing, obstacle avoidance, and navigation \cite{yu2020building}. However, the main core difference between AD and AR is the power capacity. With limited power and resources, lightweight AI systems should improvise to balance the issue of efficiency-performance tradeoff. Moreover, the price of micromobility devices is very affordable compared to IVs, which is the main factor contributing to their exponential growth in urban cities. This introduces additional complexities to the development of ARs, as it requires a quick response from authorities and research communities to maintain road safety, including pedestrians, cars, and infrastructure. However, the achievements of AD can be used to drive the adoption of micromobility growth in the form of lightweight and intelligent AR.

In the literature, few studies have been introduced as assistive technologies for the safety of two-wheelers. For example, Savino \etal highlighted a wide range of assistive technologies for powered two-wheelers that aimed to reduce crashes and improve riders' safety rates \cite{savino2020active}. Such technologies mainly provide support for two wheels to maintain vehicle control and stability such as antilock brake systems (ABS) \cite{gail2009anti}, autonomous emergency brakes (AEB) \cite{marra2021future}, traction control \cite{sarkar1998traction}, and stability control \cite{deepan2021dynamic}. ABS has been shown to reduce crash rates and improve stability during hard braking, while AEB helps to lower impact speed in unavoidable collisions. In addition, curve warning systems and haptic feedback devices improve the rider response, which may help improve riders' safety. However, these technologies represent primitive and immature in the field of autonomous riding with current challenges (see Figure \ref{fig:challenges}).

The primary goal of this survey is to bridge a critical gap in the literature by providing the first survey of autonomous riding and two-wheeler from the perspective of computer vision, sensors, and deep learning \cite{hassanin2024visual}. Although autonomous driving has been extensively surveyed, from perception and sensor fusion \cite{xiang2023multi}, motion planning and control \cite{paden2016survey}, and deep reinforcement learning frameworks \cite{sallab2017deep}, this is the first survey to present an in-depth overview of autonomous riding and its challenges. By aligning our discussion with key AD paradigms, we not only organise and compare disparate research efforts but also facilitate technology transfer from four-wheeler to two-wheeler domains. Moreover, we highlight the inherent risks and vulnerability of two-wheelers amongst road users to draw industrial relevance and commercial attention. Finally, by mapping open challenges to current needs, this survey provides a clear outlook for future work, ensuring the development of safe and scalable autonomous two-wheeler systems.
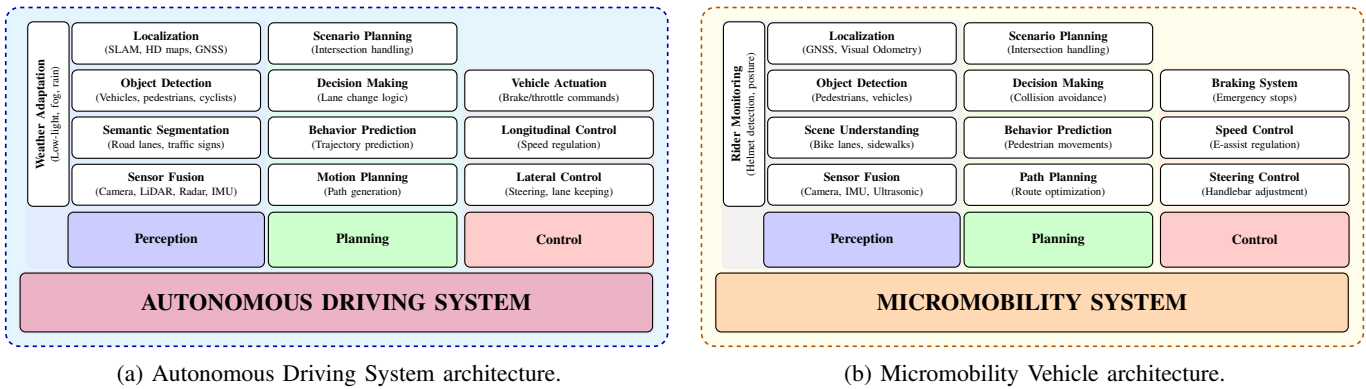
\begin{figure*}[!t]
\centering

\begin{subfigure}[t]{0.49\textwidth}
\centering
\resizebox{\linewidth}{!}{%
\begin{tikzpicture}[
    module/.style={draw, thick, rounded corners, align=center, inner sep=6pt, font=\bfseries},
    section/.style={module, fill=gray!15, minimum width=4.8cm, minimum height=1.4cm},
    subsection/.style={module, fill=white, minimum width=4.8cm, minimum height=1cm, font=\small},
    verticalsubsection/.style={module, fill=white, minimum height=1cm, minimum width=4.7cm, font=\small, rotate=90, anchor=center},
    layer/.style={fill opacity=0.3, rounded corners, inner sep=1pt}
]

\node[section, fill=purple!30, minimum width=16.1cm, minimum height=1.5cm] (ad) at (-.67,-4.6) {\LARGE \textbf{AUTONOMOUS DRIVING SYSTEM}};

\node[module, fill=blue!20, minimum width=5cm, minimum height=1.4cm] (perception) at (-5,-3) {\textbf{Perception}};
\node[section, fill=green!20] (planning) at (0,-3) {\textbf{Planning}};
\node[section, fill=red!20] (control) at (5,-3) {\textbf{Control}};

\node[subsection] (per_fusion)     at ($(perception)+(0,1.4)$) {\textbf{Sensor Fusion}\\\footnotesize(Camera, LiDAR, Radar, IMU)};
\node[subsection] (per_seg)       at ($(per_fusion)+(0,1.2)$) {\textbf{Semantic Segmentation}\\\footnotesize(Road lanes, traffic signs)};
\node[subsection] (per_det)       at ($(per_seg)+(0,1.2)$) {\textbf{Object Detection}\\\footnotesize(Vehicles, pedestrians, cyclists)};
\node[subsection] (per_loc)       at ($(per_det)+(0,1.2)$) {\textbf{Localization}\\\footnotesize(SLAM, HD maps, GNSS)};

\node[verticalsubsection] (per_weather) at ($(per_fusion)+(-3.0,1.8)$) {\textbf{Weather Adaptation}\\\footnotesize(Low-light, fog, rain)};

\node[subsection] (plan_motion)   at ($(planning)+(0,1.4)$) {\textbf{Motion Planning}\\\footnotesize(Path generation)};
\node[subsection] (plan_behav)    at ($(plan_motion)+(0,1.2)$) {\textbf{Behavior Prediction}\\\footnotesize(Trajectory prediction)};
\node[subsection] (plan_decision) at ($(plan_behav)+(0,1.2)$) {\textbf{Decision Making}\\\footnotesize(Lane change logic)};
\node[subsection] (plan_scenario) at ($(plan_decision)+(0,1.2)$) {\textbf{Scenario Planning}\\\footnotesize(Intersection handling)};

\node[subsection] (ctrl_lateral)  at ($(control)+(0,1.4)$) {\textbf{Lateral Control}\\\footnotesize(Steering, lane keeping)};
\node[subsection] (ctrl_long)     at ($(ctrl_lateral)+(0,1.2)$) {\textbf{Longitudinal Control}\\\footnotesize(Speed regulation)};
\node[subsection] (ctrl_act)      at ($(ctrl_long)+(0,1.2)$) {\textbf{Vehicle Actuation}\\\footnotesize(Brake/throttle commands)};

\begin{scope}[on background layer]
    \node[layer, fill=cyan!30, draw=blue!70!black, dashed, line width=1.2pt, rounded corners=10pt, inner sep=1em,
          fit=(perception)(planning)(control)(per_fusion)(per_seg)(per_det)(per_loc)(per_weather)
              (plan_motion)(plan_behav)(plan_decision)(plan_scenario)
              (ctrl_lateral)(ctrl_long)(ctrl_act)(ad)] {};
    \node[layer, fill=blue!15, fit=(perception)(per_fusion)(per_seg)(per_det)(per_loc)(per_weather)] {};
    \node[layer, fill=green!15, fit=(planning)(plan_motion)(plan_behav)(plan_decision)(plan_scenario)] {};
    \node[layer, fill=red!15, fit=(control)(ctrl_lateral)(ctrl_long)(ctrl_act)] {};
    \node[layer, fill=purple!20, fit=(ad)] {};
\end{scope}
\end{tikzpicture}
}
\caption{Autonomous Driving System architecture.}
\end{subfigure}
\hfill
\begin{subfigure}[t]{0.49\textwidth}
\centering
\resizebox{\linewidth}{!}{%
\begin{tikzpicture}[
    module/.style={draw, thick, rounded corners, align=center, inner sep=6pt, font=\bfseries},
    section/.style={module, fill=gray!15, minimum width=4.8cm, minimum height=1.4cm},
    subsection/.style={module, fill=white, minimum width=4.8cm, minimum height=1cm, font=\small},
    verticalsubsection/.style={module, fill=white, minimum height=1cm, minimum width=4.7cm, font=\small, rotate=90, anchor=center},
    layer/.style={fill opacity=0.3, rounded corners, inner sep=1pt}
]

\node[section, fill=orange!30, minimum width=16.1cm, minimum height=1.5cm] (mm) at (-.67,-4.6) {\LARGE \textbf{MICROMOBILITY SYSTEM}};

\node[module, fill=blue!20, minimum width=5cm, minimum height=1.4cm] (perception) at (-5,-3) {\textbf{Perception}};
\node[section, fill=green!20] (planning) at (0,-3) {\textbf{Planning}};
\node[section, fill=red!20] (control) at (5,-3) {\textbf{Control}};

\node[subsection] (per_fusion)     at ($(perception)+(0,1.4)$) {\textbf{Sensor Fusion}\\\footnotesize(Camera, IMU, Ultrasonic)};
\node[subsection] (per_seg)       at ($(per_fusion)+(0,1.2)$) {\textbf{Scene Understanding}\\\footnotesize(Bike lanes, sidewalks)};
\node[subsection] (per_det)       at ($(per_seg)+(0,1.2)$) {\textbf{Object Detection}\\\footnotesize(Pedestrians, vehicles)};
\node[subsection] (per_loc)       at ($(per_det)+(0,1.2)$) {\textbf{Localization}\\\footnotesize(GNSS, Visual Odometry)};

\node[verticalsubsection] (per_rider) at ($(per_fusion)+(-3.0,1.8)$) {\textbf{Rider Monitoring}\\\footnotesize(Helmet detection, posture)};

\node[subsection] (plan_motion)   at ($(planning)+(0,1.4)$) {\textbf{Path Planning}\\\footnotesize(Route optimization)};
\node[subsection] (plan_behav)    at ($(plan_motion)+(0,1.2)$) {\textbf{Behavior Prediction}\\\footnotesize(Pedestrian movements)};
\node[subsection] (plan_decision) at ($(plan_behav)+(0,1.2)$) {\textbf{Decision Making}\\\footnotesize(Collision avoidance)};
\node[subsection] (plan_scenario) at ($(plan_decision)+(0,1.2)$) {\textbf{Scenario Planning}\\\footnotesize(Intersection handling)};

\node[subsection] (ctrl_steering)  at ($(control)+(0,1.4)$) {\textbf{Steering Control}\\\footnotesize(Handlebar adjustment)};
\node[subsection] (ctrl_speed)     at ($(ctrl_steering)+(0,1.2)$) {\textbf{Speed Control}\\\footnotesize(E-assist regulation)};
\node[subsection] (ctrl_braking)   at ($(ctrl_speed)+(0,1.2)$) {\textbf{Braking System}\\\footnotesize(Emergency stops)};

\begin{scope}[on background layer]
    \node[layer, fill=yellow!30, draw=orange!70!black, dashed, line width=1.2pt, rounded corners=10pt, inner sep=1em,
          fit=(perception)(planning)(control)(per_fusion)(per_seg)(per_det)(per_loc)(per_rider)
              (plan_motion)(plan_behav)(plan_decision)(plan_scenario)
              (ctrl_steering)(ctrl_speed)(ctrl_braking)(mm)] {};
    \node[layer, fill=blue!15, fit=(perception)(per_fusion)(per_seg)(per_det)(per_loc)(per_rider)] {};
    \node[layer, fill=green!15, fit=(planning)(plan_motion)(plan_behav)(plan_decision)(plan_scenario)] {};
    \node[layer, fill=red!15, fit=(control)(ctrl_steering)(ctrl_speed)(ctrl_braking)] {};
    \node[layer, fill=orange!20, fit=(mm)] {};
\end{scope}
\end{tikzpicture}
}
\caption{Micromobility Vehicle architecture.}
\end{subfigure}

\caption{Comparative System Architectures for Autonomous Vehicles
(a) Autonomous Driving System: Traditional vehicle architecture featuring multi-modal sensor fusion (LiDAR/Radar/Camera), detailed environment modeling, and precise vehicle control systems. (b) Micromobility System: Lightweight urban mobility architecture emphasizing compact sensor suites (Camera/IMU), rider-aware perception, and human-centric control for bicycles/scooters. Both maintain the core perception-planning-control pipeline but differ in sensor complexity, operational domains (roads vs. bike lanes), and safety considerations (vehicle dynamics vs. rider behavior)."}
\label{fig:ad-vs-mm}
\end{figure*}
The survey is organised as follows. First, an introduction outlines the motivation and potential impact of micromobility in the context of autonomous vehicles. The background covers the difference between autonomous riding and autonomous driving in terms of technologies and visual paradigms. Following this, perception methods are discussed. Planning methods, including path prediction and activity understanding, are presented with their significance in the whole AR literature, as well as control methods. Inside each part and section, a discussion about synthesising AR from AD and future directions is provided. Two more related sections discuss datasets and applications compared to AD, which are given. Finally, the survey concludes with discussions and future directions, research opportunities, and emerging trends in the field, before concluding with the key insights of the whole survey. As shown, Figure \ref{fig:taxs} visualises the structure of AR.

The contributions of this work are summarised as follows:
\begin{itemize}
\item The first survey to provide a comprehensive overview of micromobility methods from the computer vision, sensing, and deep learning perspectives.
\item Providing a categorisation of autonomous riding and two-wheeler methods.
\item Surveying all the methods of micromobility riding.
\item Discussing the applications, challenges, and future directions of micromobility methods.
\end{itemize}

\section{Background}
\subsection{Overview of Autonomous Riding}
Autonomous riding is an emergent topic as a new paradigm, combining several technologies together, including sensors, computer vision, telecommunication, and autonomous vehicles. In this section, we cover the key aspects of other technologies that will help shape the field of visual micromobility in the future. Figure \ref{fig:ad-vs-mm} visually compares micromobility to autonomous driving as follows: (a) a conventional Autonomous Driving System (ADS) for vehicles and (b) a simplified Micromobility System for e-scooters.  To enable high-speed navigation and accurate obstacle avoidance, the ADS uses a comprehensive semantic mapping and a multi-modal sensor suite (LiDAR, radar, and stereo vision).  The Micromobility System, on the other hand, chooses a lightweight camera, IMU setup, prioritising lower costs, power usage, and vehicle mass over mapping accuracy.  Instead of modelling the entire world, its perception module concentrates on rider-aware scene knowledge (such as curb detection and pedestrian intent).  In a similar vein, micromobility design places more emphasis on rider comfort and behaviour prediction in shared urban environments than ADS planning, which stresses dynamic vehicle trajectories and high-speed stability.  Lastly, a micromobility controller delivers assistive braking that coordinates with the rider's input, whereas the control stage in an automobile ADS carries out precise throttle, brake, and steering orders.  This comparison demonstrates how micromobility platforms can enable safe, effective, and user-friendly autonomous support for two-wheelers by balancing hardware complexity with computer vision.

\subsection{Key Technologies}
\noindent\textbf{Sensors:}As micromobility continues to reshape how people move through cities, particularly with e-scooters or e-bikes, the role of sensors and onboard hardware is becoming increasingly crucial. From improving safety to enabling smart features, modern micromobility devices are now more than just simple modes of transport; rather, they necessitate intelligent machines on two wheels. Micromobility vehicles are typically smaller, energy-constrained, and less expensive than cars. This requires compact and efficient hardware settings, a balance between functionality and affordability. In this part, we summarize the sensors used in the literature.

\noindent\textbf{Camera}
Using cameras has become an integral part of our daily life applications, including surveillance and self-driving cars \cite{janai2020computer}. Cameras are passive sensors that detect colour information without emitting signals. This makes them ideal for micromobility applications due to their power efficiency. Their colour perception capability is particularly valuable for recognizing traffic lights, lane markings, and road signs. An additional advantage of the passive nature of cameras is that they avoid interference with other onboard sensors. However, cameras face significant challenges for micromobility applications, including changes in illumination and night vision. Recent advances in micromobility riding have used cameras to ensure safety for riders \cite{kaundanya2024using}.

\noindent\textbf{Fisheye Camera}
Fisheye cameras are equipped with ultra-wide-angle lenses that can capture a hemispherical field of view typically up to 180 degrees in a single frame. Unlike standard cameras that focus narrowly on what is directly ahead, fisheye lenses allow a device to include more context in the view. For micromobility, fisheye cameras are crucial where visibility is limited, and quick decisions are critical for safety; particularly, they are small size. However, more work is needed to compensate for the inherited visual distortion.

\noindent\textbf{Stereo Camera}
By using two lenses spaced a few inches apart, similar to human eyes, a stereo camera creates real 3D vision. Incorporating them in e-scooters enables them to have depth perception. This lets small vehicles judge distances to potholes, pedestrians, and curbs. Unlike single cameras that estimate depth by reconstruction, stereo pairs measure it geometrically. This makes them more reliable for such decisions needed in two-wheelers lanes and crowded streets.

\noindent\textbf{Radar}
Radio Detection and Ranging (Radar) is a sensing technology that uses radio waves to detect objects and measure their distance, speed, and direction. It exhibits robust performance in poor visibility conditions, including rain, fog, dust, and low-light environments. For micromobility, radar is not only compact, low-power, and relatively inexpensive but can also play a crucial role in rear-object detection, blind spot monitoring, and collision avoidance. Due to its high efficiency, it is necessary to provide prompt reactions for autonomous systems to avoid accidents. However, radar does not provide the same level of detail as a camera.
\begin{figure}
\centering
\scalebox{0.8}{
\includegraphics[width=\linewidth]{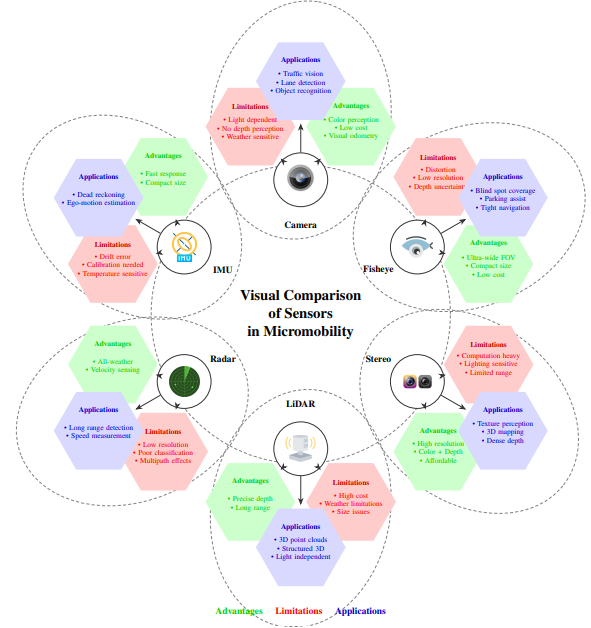}
}
\caption{Comparative analysis of sensing modalities for micromobility applications. The radial diagram evaluates six sensor types (Camera, Fisheye, Stereo, LiDAR, Radar, IMU) across key characteristics: depth resolution, environmental robustness, computational requirements, and application suitability.}
\label{fig:sensors}
\end{figure}

\noindent\textbf{LiDAR}
Light Detection and Ranging (LiDAR) is a sensing technology that utilises laser pulses to measure distances with high precision, generating cloud points for 3D processing.  In micromobility, LiDAR enables high-resolution depth perception in real time, which is important for navigating busy urban environments occluded with obstacles such as pedestrians, parked cars, curbs, and uneven roads. Although LiDAR sensors are expensive for small vehicles, some of the recent versions are compact, lightweight, and more affordable, such as Livox Mid-40 \cite{livox_mid40}. 

\noindent\textbf{IMU}
An Inertial Measurement Unit (IMU) is a compact sensor that tracks movement by measuring acceleration, rotation, and sometimes magnetic orientation. It usually combines a gyroscope, accelerometer, and occasionally a magnetometer in a single small package \cite{titterton2004strapdown}. For micromobility, IMUs are essential to provide a sense of motion, balance, and direction. Unlike other sensors, IMUs tell the vehicle about itself acceleration, whether it’s tilting, turning, or falling. This makes them useful for understanding riders' behaviours. IMUs are also so important to replace GPS when necessary. Their power efficiency, compact size, and high-speed data output signify their efficiency for compact electric vehicles.

\noindent\textbf{Ultrasonic sensor}
Ultrasonic sensors are small, low-cost devices that utilise high-frequency sound waves to detect objects and measure distance using echolocation, similar to how bats use echolocation. It provides an effective way to detect nearby obstacles in the short range.  For micromobility, they are small enough to fit easily on the front, rear, or sides with an affordable cost and weight. It is an integral to other sensors to overcome issues including low-light or adverse weather conditions.

\noindent\textbf{Pressure Sensor}
Pressure sensors are devices that detect the force applied to a surface. In micromobility, these sensors are gaining popularity for monitoring rider interaction and enhancing safety. Similar to IMU, they are important for the rider and the vehicle. For example, using it in the footboard can help detect whether a rider is properly positioned. Not only are these sensors small, low-power, and easy to integrate into two-wheelers, but they also add an extra layer of intelligence to safety systems.

\noindent\textbf{Single-beam Laser}
Single-beam laser sensors are simple yet effective devices that use a narrow laser beam to measure the distance to nearby objects. Similar to LiDAR, they emit a laser pulse and calculate how long it takes to reflect back, but in a limited scope. Additionally, it differs from LiDAR in that it scans in one direction or point at a time. In micromobility, they are low-cost, small in size, and have a fast response time.  For instance, a single-beam laser can be mounted on a stepper motor to scan a small area by rotating the beam around, allowing the system to gather depth information from multiple angles while keeping the sensor at low power consumption. 

Finally, each of these sensors has pros and cons; however, integrating them together enables the system to provide a complete and reliable safety system for two-wheelers. Table \ref{tab:sensors} and Figure \ref{fig:sensors} discuss these sensors and their usages.

\begin{table*}[!]
\centering
\caption{Overview of sensor types commonly used in autonomous micromobility research. It compares costs, operational range, accuracy, and physical size. These sensors are employed across various modules such as perception, localization, control, and safety to enhance visual processing in autonomous riding.}
\label{tab:sensors}
\begin{tabular}{l|c|c|c|c|l}
\hline
\textbf{Sensor Type} & \textbf{Cost} & \textbf{Range} & \textbf{Accuracy} & \textbf{Size} & \textbf{Methods} \\ \hline
Stereo Camera & Medium & 0.1-10m & ~82\% (object det.) & Compact & \cite{zheng2025real} \\ 
LiDAR & High & 1-200m & High (mm-level) & Varies & \cite{prabu2022wearable}, \cite{liu2018enabling} \\ 
IMU & Low & N/A & High (motion) & Small & \cite{raheel2023motorbike}, \cite{alwin2024irider} \\ 
GPS Module & Medium & Global & 1-5m (urban) & Small & \cite{poojari2024outdoor}, \cite{alai2025smart} \\ 
Ultrasonic Sensor & Low & 0.2-5m & Moderate (cm) & Small & \cite{alai2025smart} \\ 
Radar & Medium & 10-200m & High (velocity) & Medium & \cite{alai2025smart} \\ 
Pressure Sensors & Low & N/A & High (force) & Small & \cite{alwin2024irider} \\ 
Fisheye Camera & Medium & Wide FOV & Moderate & Small & \cite{Luo_2024_CVPR} \\ 
Single-beam Laser & Low & 0.1-50m & High (mm) & Small & \cite{alai2023rear} \\ \hline
\end{tabular}
\label{tab:sensors}
\end{table*}

\subsection{Applications}
Because of the compact size and affordable prices of e-bikes and two-wheelers, AV industry will be reshaped with various applications. Here, we are listing some of the applications that are going to emerge in the near future.
\begin{itemize}
    \item \textbf{Autonomous Driving:}  
      From a high-level view, micromobility can be part of autonomous driving, so integrating them into the broader domain of AVs will benefit both parties. For instance, standard benchmarks like KITTI and nuScenes include up to 12\% two‐wheeler instances, leading to false negatives in e-scooters and motorcycles detection \cite{geiger2012kitti} \cite{caesar2020nuscenes}.  From the perspective of AVs, the widespread use of two-wheelers on the road will create a crisis for them and dictate a quick response to ensure the safety of both vehicles and riders. In other words, the progress of either two-wheelers or AVs will surely benefit the other, as the relationship is mutually beneficial. 
    
    \item \textbf{Rider Safety Systems:}  
      Rider safety is a crucial matter, not only for himself but also for all the road users, including the infrastructure. Integrating with all the legacy systems and road sensors is essential to ensure the complete safety of the system.
    
    \item \textbf{Smart City Infrastructure:}  
      Vision‐based analytics of two‐wheeler vehicles is a fundamental aspect of a smart city. Continuous monitoring of two-wheelers' hotspots and violation events allows cities to dynamically adjust signal timings, allocate micromobility lanes, and dispatch enforcement resources where riders exist.

    \item \textbf{Urban Planning and Policy:}  
      Aggregating large‐scale two‐wheeler trajectory data reveals preferred routes for two-wheelers, especially with the widespread use of such vehicles inside crowded cities and capitals.  Planners can use these insights to justify dedicated micromobility lanes, set evidence‐based regulations, and hence balance safety, accessibility, and traffic efficiency.
\end{itemize}

\section{Perception}
Different from conventional autonomous cars, perception for autonomous riding encounters several challenges.  
When paired with congested urban environments, the dynamic behaviour of e-scooters and e-bikes produces complicated perception situations that result in occlusions, quick motions, and variable illumination conditions.  Three crucial areas have advanced the realms of perception in autonomous riding through recent developments in edge computing and computer vision: violation detection, re-identification of e-bikes, and rider tracking.  However, there are still many restrictions when it comes to addressing edge scenarios, especially when it comes to crowded settings, low-light levels, and obstructed riders. This section provides a roadmap for future research by reviewing the state-of-the-art in micromobility perception and analyzing current gaps within each paradigm. Discussion about how to progress the field of autonomous riding to catch up with the advancements of autonomous driving is provided at the end of each subsection.

\subsection{Rider Detection and Tracking}
The increasing popularity of two-wheeler riding, including from motorcycles to e-scooters, has brought forth significant challenges in traffic safety and regulation enforcement \cite{nayak2023advances}. One of these challenges is the need to accurately detect the presence of riders in real-time, particularly with all the environmental changes, such as night vision, adverse weather, and occlusion. Autonomous Rider Detection aims to leverage advancements in computer vision, edge AI, and sensing to detect riders. Such models are crucial for maintaining riders' safety within the whole driving system. This section explores the key technologies used in autonomous rider detection, including object detection models and deep learning models. 
Using edge AI platforms, including NVIDIA Jetson and Raspberry Pi, \cite{howard2017mobilenets} proposed to detect motorbikes in live video feeds with low latency in smart traffic monitoring, urban surveillance, and intelligent transportation systems. Model compression techniques, including pruning and quantization, and use edge inference optimizations such as TensorRT or OpenVINO have been applied. \cite{apurv2021detection} proposed a very basic study to differentiate between e-scooter riders and pedestrians in natural scenes. A custom dataset is collected to provide a benchmark with the recent CNN algorithms, including MobileNetV2 \cite{sandler2018mobilenetv} and YoloV3 \cite{redmon2018yolov3}. \cite{lee2025yolov5} combined YOLOv5 \cite{zhang2022electric} object detection model with GPS data, Raspberry Pi 5 hardware, and Amazon Web Services (AWS) for web-based implementation and data processing.  Two YOLOv5 variants were used, one for detecting e-scooters and the other for helmetless riders or triplet riders. \cite{gilroy2022scooter} proposed a novel approach to detect e-scooter riders in dense urban environments that have technical challenges due to complex backgrounds, occlusions, and diverse riding behaviors. This work collected a dataset with 1130 images of persons and e-scooters, along with pose estimation and motion pattern analysis. This helped to improve the accuracy compared to the related works, and it achieved acceptable results in safety monitoring. However, limitations in handling severe occlusions remain.  \cite{sabri2024detection} proposed a novel method to detect micromobility vehicles (e-scooters, e-bikes, etc.) in urban traffic videos. A feature maps aggregation of consecutive frames is proposed to YOLO's architecture to improve the understanding of urban mobility patterns. This helped address the inherited challenges from the traditional scenes, especially in the absence of the related datasets.  \cite{kim2024development} proposed a novel system to recognize the e-scooters with fine details using YOLO \cite{redmon2016you}.  Chen \etal provided a comprehensive study on performance evaluation on e-scooter detections using various versions of YOLO \cite{chen2024performance}.

Finally, rider and e-scooter detection have been neglected by the research community, as the number of papers published within the last three years does not exceed ten. Moreover, the work is nearly centered around using YOLO-like models for detection. This reflects the weak attention for such important paradigms, particularly since it is very crucial for maintaining riders' safety in such a complex environment. Here, we highlight the open gaps related to this section to draw the attention of researchers.

\noindent\textbf{Gaps and Directions inspirations from autonomous driving}

\noindent Recent advancements in AVs have significantly enhanced their ability to perceive, interpret, and navigate complex environments. A foundational progression in AVs is the implementation of sensor fusion, which combines data from LiDAR, radar, and cameras to create a comprehensive understanding of the vehicle's surroundings. This integration allowed for accurate object detection even under challenging conditions such as low light or adverse weather. Several approaches have been proposed in the literature, including real-time object detection using YOLOs \cite{zuraimi2021vehicle}, the use of Large Language Models (LLMs) for contextual reasoning \cite{karagounis2024leveraging}. LLMs can interpret complex scenarios by understanding the relationships between various elements in the environment. Another factor of AVs' progressions is the development of Vehicle-to-Everything (V2X) communication that allows AVs to share sensory data with each other \cite{anaya2015vulnerable}. This extends their awareness beyond and builds towards a comprehensive understanding of the whole road. Other road participants' behaviors have also been discussed in the literature by incorporating models that account for the intentions and vulnerabilities of pedestrians, cyclists, and other drivers \cite{wang2020vulnerability}. Sensor fusion is present significantly in the table of AVs to help overcome occlusion and clutter \cite{zhang2008multilevel}.

\noindent Improving micromobility detection and tracking within autonomous vehicle (AV) systems is a mutual need for vendors and riders. Micromobility users, such as e-scooter and e-bike riders, present unique challenges due to their small size, rapid movements, and persistent occlusions in the cities. However, several research directions and technological solutions are emerging to address these challenges effectively. One promising approach involves the integration of spatio-temporal detection models that leverage both spatial and temporal information to enhance detection accuracy. For instance, aggregated feature maps from consecutive frames to process will definitely be better than still images. Research has achieved substantial improvements in detecting tiny objects in various paradigms, which can be easily applied to micromobility \cite{coluccia2021drone, coluccia2021drone}. Sensor fusion techniques are also critical in this context. Combining data from multiple sensors, such as cameras, LiDAR, and radar, can compensate for the limitations of visual cues. For example, fusing LiDAR and camera data enables more accurate detection and tracking of micromobility users, even in complex scenes. Lightweight models should be used to process data generated by these sensors in real-time. Recent studies have demonstrated the effectiveness of edge-based computing systems in achieving high accuracy and low latency in object detection tasks, and hence, adaptation to micromobility is necessary.
    
\subsection{Re-Identification and Counting:}
\noindent \textbf{Motorcycles Re-Identification}\cite{figueiredo2021more} collected the first large-scale Motorcycles Re-Identification (MoRe) dataset with 3,827 individuals (\ie the set of motorbikes and motorcyclists) captured by ten surveillance cameras. Furthermore, a deep learning model trained for object re-identification provides a baseline for the Motorcycles MoRe problem. 

\noindent\textbf{Micromobility counting:} \cite{huynh2021motorbike} addressed the challenge of accurately counting motorbikes in dense traffic environments. Traditional object detection and tracking methods often struggle in such scenarios due to heavy occlusion, scale variation, and dense object distribution. To overcome these limitations, a deep learning-based crowd counting approach adapted from people counting in dense crowds is used. A performance evaluation has been applied to various deep learning techniques to illustrate the position of such a problem in the literature.

\noindent\textbf{Gaps and Directions inspirations from autonomous driving}
For autonomous driving, vehicle re-identification (Re-ID) and vehicle counting have been studied to improve situational awareness and traffic analysis. Addressing the challenge, including varying viewpoints, occlusions, and environmental conditions, has been prioritised in the literature. One of the directions involved the use of the most recent techniques in Person ReIDs and machine learning \cite{zheng2016person}. For instance, the IBNT-Net model combines a ResNet50-IBN backbone with an improved multi-head self-attention mechanism \cite{wang2025vehicle}. This approach addresses intra-class variations and inter-class similarities to increase the discrepancy between the candidates, and hence, better performance. Self-supervised learning techniques have been utilised for vehicle re-identification to reduce the reliance on labeled datasets \cite{khorramshahi2023robust}. Such approaches have shown potential in improving scalability and adaptability across diverse environments. Several methods have been developed with the aim of re-identifying vehicles under various environmental conditions, including day/night vision \cite{li2024day}, attention-based \cite{hu2024tanet}, language-based \cite{wang2024vision}, and domain adaptation \cite{xu2025clip}.

In the context of vehicle counting, progress has been made through the development of end-to-end deep networks that combine object detection, tracking, and counting. Such systems utilize YOLOv4 for real-time object detection, combined with tracking algorithms to accurately count vehicles in dynamic traffic scenes \cite{albouchi2024implementation}. As many approaches have been suggested in the literature for that purpose, for example, video-based counting \cite{dai2019video}, overlapping vehicles \cite{guerrero2015extremely}, from aerial videos \cite{xiang2018vehicle}, and sensor-based counting \cite{taghvaeeyan2013portable}. 

With uniform appearance, absence of license plates, compact size, and frequent occlusion in urban environments, advancing micromobility counting and re-identification is very challenging. Special algorithms need to be developed with occlusion-aware micromobility. Joint models are also a promising strategy to distinguish between riders and vehicles. Another cues from the environment need to be utilized, such as place ID, or geography coordinates, to build discriminated patterns. Occlusion-aware methods in vehicle re-identification are another possible alternative, particularly when combining visual data with sensor inputs such as LiDAR and GPS.
Additionally, using synthetic data to train models can help address the scarcity of labeled datasets for micromobility vehicles. Furthermore, integrating temporal information through video sequences allows for better tracking and re-identification. Finally, domain adaptation and transfer learning methods remain one of the best solutions for projecting the advancements of autonomous driving into autonomous riding, particularly in zero-shot learning. This can be advanced more rapidly after the introduction of visual-language models such as CLIP \cite{radford2021learning} and DINO \cite{caron2021emerging}.

\begin{table*}[!h]
\centering
\caption{Visual summary of perception challenges, technical approaches, sensor modalities, and research gaps in Autonomous Micromobility. This table categorizes core perception capabilities including rider detection, re-identification, and violation detection, along with current solutions and future directions inspired by autonomous driving.}

\label{tab:perception_summary}
\renewcommand{\arraystretch}{1.2}
\resizebox{\textwidth}{!}{%
\begin{tabular}{|l|l|l|l|l|}
\hline
\textbf{Component} & \textbf{Key Challenges} & \textbf{Techniques/Models} & \textbf{Sensors} & \textbf{Gaps \& Future Directions} \\ \hline

\textbf{Rider Detection \& Tracking} & 
\begin{tabular}{@{}l@{}}
• Severe urban occlusions \\
• Small object size \\
• Limited datasets \\
• Night/low-light conditions
\end{tabular} & 
\begin{tabular}{@{}l@{}}
• YOLO variants (v3-v8) \\
• MobileNet edge deployment \\
• Feature map aggregation \\
• Pose estimation
\end{tabular} & 
\begin{tabular}{@{}l@{}}
• RGB cameras \\
• Edge devices (Jetson/RPi) \\
• GPS/IMU fusion \\
• AWS cloud processing
\end{tabular} & 
\begin{tabular}{@{}l@{}}
• Occlusion-aware architectures \\
• LiDAR/radar fusion \\
• Synthetic data generation \\
• Standardized benchmarks
\end{tabular} \\ \hline

\textbf{Re-Identification \& Counting} & 
\begin{tabular}{@{}l@{}}
• No license plates \\
• Uniform appearance \\
• Viewpoint variations \\
• Dense urban scenes
\end{tabular} & 
\begin{tabular}{@{}l@{}}
• MoRe dataset baseline \\
• Crowd-counting adaptations \\
• Spatio-temporal models \\
• Attention mechanisms
\end{tabular} & 
\begin{tabular}{@{}l@{}}
• Surveillance cameras \\
• Multi-view systems \\
• Aerial cameras \\
• Embedded sensors
\end{tabular} & 
\begin{tabular}{@{}l@{}}
• Visual-language models \\
• Occlusion-aware Re-ID \\
• Domain adaptation \\
• V2X integration
\end{tabular} \\ \hline

\textbf{Violation \& Anomaly Detection} & 
\begin{tabular}{@{}l@{}}
• Real-time edge constraints \\
• Data imbalance \\
• Fisheye distortions \\
• Low-light conditions
\end{tabular} & 
\begin{tabular}{@{}l@{}}
• Ensemble detectors \\
• Deep SORT tracking \\
• Coarse-to-fine detection \\
• Test-time augmentation
\end{tabular} & 
\begin{tabular}{@{}l@{}}
• Traffic cameras \\
• Dashcams \\
• Fisheye cameras \\
• Wearable sensors
\end{tabular} & 
\begin{tabular}{@{}l@{}}
• IMU-vibration pattern analysis \\
• Edge TPU optimization \\
• Wearable intoxication screening \\
• Automated e-ticketing
\end{tabular} \\ \hline
\end{tabular}%
}
\label{tab:perception_summary}
\end{table*}
\subsection{Violation and Anomaly Detection}
This section covers vision-based detection of traffic violations, including helmetless riders, e-scooter sidewalk riding, signal jumping, and speed violations. Emphasis is placed on real-time, lightweight methods suitable for edge devices deployed on roads, vehicles, or rider helmets. Relatively, this section is considered the most popular part among researchers as it includes more papers in the field.

\noindent\textbf{Helmet Detection and Tracking:} 
Hossain \etal collected a dataset of bikes and used object detectors such as SSD \cite{liu2016ssd} and Faster-RCNN \cite{girshick2015fast} for detecting helmetless riders \cite{hossain2021identifying}. They also recognized that plate numbers are using Tesseract-OCR \cite{smith2007overview} for tracking reasons. \cite{tran2023robust} introduced detecting motorcycle helmet violations in real-time to enhance road safety. YOLO v8 \cite{jocher2020ultralytics} has been used for the detection of motorcycles in traffic cameras. Akhtar \etal \cite{akhtar2024real} proposed a lightweight and efficient motorbike detection system suitable for real-time applications on edge devices. Unlike traditional methods that require high computational resources, the proposed approach processes deep learning models, such as  YOLO and MobileNet.  \cite{duong2023helmet} proposed an approach for automatically identifying motorcyclists without helmets. State-of-the-art object detection models (such as YOLOv5 or SSD) with a custom tracking framework (Deep SORT \cite{wojke2017simple}) are integrated to enhance detection reliability. This integration between object detectors and trackers helped to solve the occlusion in the scenes.  \cite{hernandez2024computer} collected a dataset for pedestrian and motorcycle scenarios, more precisely, motorcycles in crosswalks, motorcycles outside crosswalks, pedestrians in crosswalks, and only motorcycles outside. CNN object detectors (YOLOv8, SSD, and MobileNet) are used to detect violations to improve riders' safety. \cite{rawat2025dashcop} introduced an end-to-end system that automates the detection of traffic violations by two-wheeler riders using dashcam footage. Traditional object detection models (\eg YOLO or SSD) are used to locate motorcyclists' helmet compliance. Also, a vehicle tracking algorithm ensures continuity across frames, and license plate recognition is applied to accurately identify violators. An e-ticket containing evidence such as video frame snapshots, violation type, timestamp, and vehicle registration details is issued once a violation is detected. \cite{goyal2022detecting} proposed using curriculum learning-based object as an object detector for helmetless riders, especially in cases of occlusions, whereas a novel trapezium-shaped object boundary representation is used to fix the rider-motorcycle association. This approach proved its superiority over the state-of-the-art. \cite{zhang2024coarse} proposed a Coarse-to-fine helmet detection: Coarse detector suggests the initial location of both riders and bikes, Fine-grained with a classification head to detect drivers from passengers. Test Time Augmentation (TTA) and Weighted Boxes Fusion (WBF) are used to augment the dataset as a way of fixing data imbalance. \cite{Luo_2024_CVPR}  proposed traffic detection for traffic monitoring using fisheye cameras. For image distortions and challenges, FE-Det integrated advanced object detectors like YOLOv8 and InternImage \cite{wang2023internimage} to handle different lighting conditions, such as day and night. Pre-processing techniques are employed to correct fisheye distortions, including rotational ones, as well as to improve localization accuracy. \cite{wang2023prb} presented a two-step approach to track helmet violations. PRB-FPN+, an object detector that excels in object localization. They also explored the benefits of deep supervision by incorporating auxiliary heads within the network.  This enhanced performance of the proposed method. Followed by object tracking to filter the drivers from the passengers.

\noindent\textbf{Detectors Ensemble:} 
\cite{van2024motorcyclist} proposed addressing the issues of varying object sizes and imbalanced datasets in motorbike helmet detection datasets.  An ensemble method is developed to benefit from various object detection models such as YOLOv7, YOLOv8, Co-DETR \cite{zong2023detrs}, and EfficientDet \cite{tan2020efficientdet}. Two copy-and-paste data augmentation techniques have been proposed to fix data imbalance in the collected dataset. A very similar study to \cite{van2024motorcyclist}, \cite{kim2024helmet} used CNN- and Transformer-based models for ensemble, while Weighted Box Fusion (WBF) \cite{solovyev2021weighted}  and Curriculum of Data Augmentation (CUDA) \cite{ahn2023cuda} are used for precise detection and augmentation, respectively.
The same problem of data imbalance is also addressed in \cite{vo2024robust} by traditional object detectors; however, Minority Optimizer and the Virtual Expander are used for striking imbalanced data.

 \noindent \textbf{Triple riding:}\cite{mallela2021detection,charran2022two, srilekha2022detection}  addressed road safety violations on two major infractions: triple riding and speed limit breaches. Deep learning techniques, including CNNs, and object tracking methods combined with calibrated distance and time measurements have been employed. \cite{bose2023loltv} introduced a novel dataset for detecting two-wheelers violations under low-light conditions, such as at night or in poor visibility. An anomaly detection framework is proposed to detect various violations, including riding without a helmet and triple riding. Such a method is crucial, especially in countries where low-light traffic monitoring is present.
   
\noindent\textbf{Gaps and Directions inspirations from autonomous driving}
Autonomous cars have matured enough to identify traffic laws. Real-time detection of stop signs, speed limits, and traffic lights is made possible by CNN-based traffic sign recognition systems, which achieve over 99\% accuracy on public datasets such as GTSRB \cite{stallkamp2012man}. More precisely, after the introduction of sensors, including HD maps and GPS data along with visual inputs to prevent violations such as running red lights \cite{n2025real, youssouf2022traffic}.  Collision-avoidance modules combine camera, radar, and LiDAR data to predict and prevent impending crashes, initiating emergency braking or steering maneuvers by detecting anomalies.

To bridge these gaps, much attention from the research community is required to produce edge-capable, lightweight, and sensing-based models. For example, a helmet-use algorithm with YOLOv3  achieved  4.4\% improvements in overall helmet detection precision after legislation \cite{siebert2023computer}. Sensor fusion of IMU, GPS, and vision data can flag sidewalk riding by connecting typical vibration patterns with mapped road segments. Edge processing on micro-GPUs or TPUs (e.g., Coral, Jetson Nano) can provide the environment for AI processing, enabling real-time alerts such as warnings or automated speed decreases when violations occur. Rider-state estimation by wearable sensors or smartphones can screen for intoxication, lowering incidences connected to alcohol by a large margin. Table \ref{tab:perception_summary} provided a visual summary of the various components, challenges, proposed methods, sensors, and future directions for perception in micromobility.

\section{Planning}
Autonomous micromobility systems should utilise observed information to provide safe, goal-oriented actions. This includes anticipating rider intent, avoiding hazards, and creating efficient routes.  E-bikes and e-scooters, in contrast to full-sized vehicles, must react to both rider actions and dynamic scenarios with less to fit the restricted settings.  In this section, we first cover techniques for understanding rider activity, which combine multimodal recordings (IMU, video, and eye-tracking) and time-series sensor fusion to forecast movements, identify distractions or fatigue, and assist in real-time collision avoidance.  Then, autonomous path navigation covers both indoor and outdoor settings, from GPS/IMU-driven route planning and live obstacle detection on city streets to stereo-vision and LiDAR-based localisation in corridors.  We demonstrate how developments in lightweight sensor fusion and sequence modelling (LSTM, Transformers) can be customised to the particular limitations of micromobility systems. Discussion about how to progress the field of autonomous riding to catch up with the advancements of autonomous driving is provided at the end of each subsection.

\subsection{Rider Activity Understanding}
Predicting rider behaviors such as turning, sudden stops, and lane weaving is critical for maintaining the safety of the riders and the road partners' safety. This section explores deep learning techniques for forecasting micromobility riders' behavior.

 \cite{zhu2020understanding} provided a study to compare between bike-sharing and scooter-sharing. It concludes that scooter sharing is more effective, with smaller fleet sizes and higher daily usage, particularly in dense areas. However, e-scooters have parking problems, as well as redistributing the e-scooters in the needed areas. 
    
 An early study of Brunner \etal for e-scooter riders' safety is provided in \cite{brunner2020analysis}. They recommended that dynamical modes, such as the control and guidance process performed by the rider of e-scooter, should be conducted to improve safety and stability for autonomous riders. \cite{raheel2023motorbike} proposed a method for recognizing motorbike-related driving activities using the motion sensors of smartphones, such as the accelerometer and gyroscope. Mainly, motorbike activities (e.g., riding, idling, turning, braking, or stopping) are identified to provide safety precautions through rider behavior analysis. Time-series sensor data collected during various driving scenarios and processes are evaluated using classifiers such as Random Forest, SVM, and deep learning to detect motorbike actions accurately. Results show high recognition accuracy, demonstrating the feasibility of using mobile sensors for real-time activity recognition of motorbikes. However, this method lacks the use of images and videos. Therefore, the dynamic environmental context is missing.  \cite{kegalle2025watch} explored the riding behaviors of autonomous riders in a naturalistic study. 23 participants were hired and equipped with a bike computer, eye-tracking glasses and cameras. Then, they traversed a pre-prepared route to collect multi-modal data. The findings of this study highlighted the challenges of autonomous riders, including issues in safety due to the risks of losing control when using hand signals, and limited acceptance from other road users in mixed-use spaces. iRider \cite{alwin2024irider} 
provided a complete system that combines cameras and sensors to evaluate electric scooter riders' biomechanics in real time. Information on rider posture, movement, and stability while using a scooter has been collected by inertial measuring units (IMUs), pressure sensors, and video input to achieve safety. \cite{prabu2022wearable} provided a wearable data gathering system to comprehend the behavior of autonomous riders. LiDAR, cameras, and GPS are combined in a system controlled by ROS. This helped draw some conclusions about the behavior of e-scooter riders. \cite{nguyen2024remote} presents a cutting-edge safety and health monitoring system that uses edge computing and sensors to assess the rider's posture and movement. A small video dataset is collected from 11 patients with precise kinematic information, such as joint angles, body alignment, and motion patterns. This data is evaluated locally using edge AI models to identify anomalies, weariness, or risky riding behaviors in real time.  This facilitates continuous mobility assessment for older or disabled users, enhances ride ergonomics, and promotes early identification of health problems. In \cite{tabatabaie2024beyond}, insights about the interactions and behaviors of autonomous riders in urban settings of human-mobility interactions (HMI) are derived. A novel method, namely E-scooter Naturalistic Riding Understanding System (CENRUS) is proposed based on sensing,
analyzing, and understanding the behavioral, visual, and textual annotation data of RXIs (Rider-to-X Interactions). The main goal of CENRUS is to serve as a foundational system for safe autonomous riding.

\noindent \textbf{Collision avoidance:}   \cite{white2023factors} has conducted a naturalistic riding study that sheds light on the causes of e-scooter crashes. Over six months, extensive data on rider behavior and ambient circumstances by outfitting a fleet of 200 e-scooters with sensors and cameras have been collected. The results show that the probability of collisions is significantly increased by rider distraction, such as distracted riding and disregard for traffic laws. Infrastructure-related conditions, such as bad road conditions and a lack of designated scooter lanes, were also among the leading causes behind the accidents. \cite{tabatabaie2023naturalistic} proposed FCRIL, a novel framework for identifying, understanding, and learning rider manoeuvres and behaviors on electric scooters in naturalistic environments.  In contrast to conventional centralized approaches, FCRIL protects user privacy by training local models that are regularly aggregated into a global model using raw sensor data (gathered from smartphone-based IMUs) on edge devices. This derives relevant rider maneuver patterns, such as time series, and other statistical features from the collected dataset, facilitating other useful applications, including safety measures and collision avoidance.
An analysis for e-scooter riders' safety is provided in \cite{brunner2020analysis}, which recommends that the guidance process performed by the rider of an e-scooter should be conducted to improve safety and stability for autonomous riders.

\noindent\textbf{Gaps and Directions inspirations from autonomous driving:}

\noindent Advanced driver assistance systems (ADAS) and contemporary autonomous driving now rely extensively on driver activity identification \cite{antony2021advanced, qu2024comprehensive}. Several approaches have been proposed to study driver behaviour in autonomous driving including fatigue and distraction \cite{craye2016multi}, EEG-based model on the SEED-VIG benchmark \cite{siddhad2025awake, zheng2017multimodal},  mouth yawns and eye closures real-time drowsiness monitoring on standard video datasets \cite{jiao2020driver}, fatigue detection under varied lighting conditions \cite{peivandi2023deep}. Also, the most recent machine learning techniques have been employed to improve the accuracy and performance, such as LSTM, Transformers \cite{vaswani2017attention, hassanin2022crossformer}, GAN-augmented LSTM \cite{jiao2020driver}, Capsule-based LSTM networks \cite{alparslan2020towards}. Fusion of visual features, including head-pose and facial features  \cite{yu2024driver}, has also been investigated. Beyond brain signals, novel approaches used various sensors including sEMG sensors in the steering wheel \cite{lu2021can}, fusion of external LiDAR point-cloud dynamics with radar Doppler profiles has also been explored to infer erratic steering and lane-change behaviors, \cite{li2024intention}. Finally, these directions show the amount of work that has been done for autonomous driving, which represents a cornerstone of micromobility.

On the other hand, bridging the gap between AD and AR involves applying the various progressions of AD in AR, considering the restrictions of two-wheelers in terms of space and computation. This presents unique challenges due to the vehicles' physical constraints and the nature of rider interaction. However, several adaptations are feasible: 1) While traditional vehicles can mount extensive sensor arrays, micromobility devices require compact solutions. One of the solutions is to utilize smartphone sensors (accelerometers, gyroscopes), which can facilitate the collection of riders' behavior data in a cheaper and easier way. 2) Using real-time feedback analysis to predict the unsafe behaviors of riders and then inform riders of such behaviors.

\subsection{Autonomous path navigation}

\noindent\textbf{Indoor Navigation:} In \cite{liu2018enabling}, Liu \etal proposed a multi-modal approach to address the need for hassle-free navigation for individuals with mobility challenges, such as disabled people. In indoor settings, the normal sensors fail to provide the correct locations, and an intelligent autonomous scooter is proposed to enhance safety where traditional navigation aids may fall short. Data from stereo vision cameras, synthetic laser scanners, and LiDAR sensors is fused for both fine-grained resolution and long-range coverage.

\begin{table*}[!h]
\centering
\caption{Visual summary of main functional components, techniques, sensors, and open challenges in Autonomous Micromobility Planning. This summary categorizes core capabilities, including rider behavior understanding, collision avoidance, and indoor/outdoor navigation. As well as the proposed models, sensing modalities, and research opportunities that can be adapted from AD.}

\label{tab:planning_summary}
\renewcommand{\arraystretch}{1.2}
\resizebox{\textwidth}{!}{%
\begin{tabular}{|l|l|l|l|l|}
\hline
\textbf{Component} & \textbf{Goal} & \textbf{Techniques/Models} & \textbf{Sensors} & \textbf{Challenges \& Future Directions} \\ \hline

\textbf{Rider Activity Understanding} & 
Predict rider behavior (turns, braking, fatigue) & 
\begin{tabular}{@{}l@{}}
• Time-series classification \\
• LSTM/Transformer models \\
• Multimodal fusion \\
• Edge AI optimization
\end{tabular} & 
\begin{tabular}{@{}l@{}}
• IMU (accelerometer/gyro) \\
• Camera vision \\
• Wearable sensors \\
• Eye-tracking
\end{tabular} & 
\begin{tabular}{@{}l@{}}
• Limited contextual awareness \\
• Need for mobile-based sensing \\
• AD-inspired fatigue detection
\end{tabular} \\ \hline

\textbf{Collision Avoidance} & 
Prevent accidents and near-misses & 
\begin{tabular}{@{}l@{}}
• Federated Learning (FCRIL) \\
• Naturalistic behavior studies \\
• Real-time prediction models
\end{tabular} & 
\begin{tabular}{@{}l@{}}
• Smartphone IMUs \\
• Onboard cameras \\
• Ultrasonic sensors
\end{tabular} & 
\begin{tabular}{@{}l@{}}
• Infrastructure hazards research gap \\
• Privacy-preserving models \\
• Multi-rider data fusion
\end{tabular} \\ \hline

\textbf{Indoor Navigation} & 
Enable precise indoor mobility & 
\begin{tabular}{@{}l@{}}
• Multi-modal SLAM \\
• Vision-LiDAR fusion \\
• Edge-optimized mapping
\end{tabular} & 
\begin{tabular}{@{}l@{}}
• Stereo cameras \\
• Compact LiDAR \\
• Depth sensors
\end{tabular} & 
\begin{tabular}{@{}l@{}}
• Sensor noise robustness \\
• Cross-domain adaptation \\
• Low-power requirements
\end{tabular} \\ \hline

\textbf{Outdoor Navigation} & 
Urban path planning and safety & 
\begin{tabular}{@{}l@{}}
• TEB local planner \\
• TrajectoFormer \\
• Graph neural networks
\end{tabular} & 
\begin{tabular}{@{}l@{}}
• GPS/IMU fusion \\
• Stereo/RGB cameras \\
• Single-beam LiDAR
\end{tabular} & 
\begin{tabular}{@{}l@{}}
• Standardization needed \\
• Dynamic obstacle handling \\
• AD-inspired fusion methods
\end{tabular} \\ \hline
\end{tabular}%
}
\label{tab:planning_summary}
\end{table*}

\noindent\textbf{Outdoor Navigation:} An Early study, a car-like drive model and Timed Elastic Band (TEB) \cite{rosmann2017kinodynamic} local planner within ROS have been used for the dynamics of the scooter and motion planning, respectively. A graph with lower edge weights for commonly traveled roads using previous ride data from Veo \cite{veo2025website} has been used to improve global path planning. \cite{mulky2018autonomous} proposed a navigation system to enhance mobility for individuals with physical disabilities. A safe and independent mobility path is provided along with voice-command assistance to help disabled riders avoid obstacles and arrive safely. Though it is a novel study and lacks the ability to overcome the complicated nature of real environments, it is still a promising study in that direction. \cite{alai2023rear} proposed a sensing system to track vehicles coming from behind to improve e-scooter rider safety. A cheap single-beam laser sensor is mounted on a stepper motor that dynamically shifts its orientation to focus on the right front corner of the car when it detects one. The system utilizes real-time trajectory estimation to alert approaching riders of a potential collision by sounding an audible alarm.  An innovative method for autonomous e-scooters throughout cities has been introduced to find their own way to parking spots without assistance from humans \cite{10711284}. A stereo camera for depth perception, IMUs, GPS modules for localization, motor drivers for control, and an Nvidia Jetson Orin and ROS have been integrated into a compact toolkit. \cite{zhang2024intent} proposed path prediction for the movements of autonomous riders to improve road safety. Egocentric and bird's-eye views collected from various urban settings across the US are used by a deep learning model to predict the trajectory based on rider behaviors and environmental settings. High accuracy in predicting the trajectories of vulnerable road users has been obtained, outperforming traditional linear models in terms of Average Displacement Error (ADE) and Final Displacement Error (FDE). This combination of deep learning and sophisticated sensing technologies has helped develop flexible, intelligent transportation systems that can adjust automatically. Poojari \etal proposed a self-driving e-scooter that is able to navigate pre-planned paths \cite{poojari2024outdoor}.  Robot Operating System
(ROS) on Ubuntu Linux is used to run the scooter along with a stereo
camera for depth sensing, an IMU, a GPS module and other tools. These readings are used to predict and optimize the paths.  However, it is limited to outdoor settings as well as it is offline which fails to handle the sudden changes. \cite{alai2025smart} proposed learning the trajectories of the vehicles behind the e-scooter to help secure the riders from collisions by anticipating the motions of cars approaching. The scooter continuously analyses its environment using a combination of GPS and inertial measurement units, as well as rear-facing sensors like cameras, ultrasonic sensors, or radar. Deep learning-based motion prediction models analyze the trajectories of nearby vehicles to anticipate potential collisions or unsafe proximity. When a risk is detected, an alert to the rider through haptic feedback, auditory warnings, or visual cues on the display is given. The scooter may turn on automated braking or emergency lights in dire circumstances. In addition to reducing the likelihood of rear-end crashes, this integration of real-time perception encourages safer micromobility in urban settings. \cite{zheng2025real} proposed a novel system to make e-scooter riding safer by detecting obstacles, particularly small bumps, potholes, and uneven surfaces. An RGB camera, a depth sensor, and an IMU sensor are integrated into a compact Intel RealSense Camera D435i. Six types of dangerous obstacles, such as tree branches, potholes, manhole covers, and cracks, are detected using both visual clues and the vibrations felt by the scooter. YOLO is used for object detection to quickly identify obstacles from the camera images, while depth data helps determine the actual distance of these hazards.  By achieving a significant accuracy (82\%) during seven hours of real-world scooter trips, the way to achieve safety for autonomous riders is paved.

\noindent\textbf{Gaps and Directions inspirations from autonomous driving}

\noindent Path prediction in autonomous driving has recently progressed from simple kinematic models to data-driven frameworks with temporal data and driver interactions \cite{rudenko2020human}. Machine learning models such as LSTM and Attention-LSTM model used connected vehicles within a specific radius to forecast vehicle paths \cite{Lin2022Attention}, Graph Neural Networks (GNNs) to encode pairwise interactions between vehicles, each vehicle as a node and their influences as edges \cite{wang2024lstm}, Graph-Transformer hybrids to fuse semantic map data and dynamic graph attention \cite{singh2022multi}, GAN-augmented LSTM models to enhance diversity by producing multiple plausible future trajectories \cite{rossi2021vehicle}, deep spectral methods to employ spectral graph convolutions and temporal gated convolutions \cite{cao2021spectral}, Transformer architectures, which use self-attention to capture long-range dependencies across agents and time steps \eg  TrajectoFormer \cite{li2024trajectory}, even  Physics-Informed Transformers (PIT-IDM) embed kinematic constraints directly into the attention framework \cite{geng2023physics}. Finally, these directions, spanning LSTM, GNNs, GANs, spectral methods, and Transformers, summarise the evolution of trajectory prediction in autonomous driving, laying a robust foundation for next-generation micromobility path planning.


Due to the nature of micromobility, sensing fusion is required to collect information to help AI predict path planning, especially since micromobility is not confined to lanes, and vehicles can drive on side roads. Therefore, integrating lightweight, high‐fidelity sensors, such as fisheye cameras, mini‐LiDAR, radar, and IMUs, is necessary to perceive the surrounding environment reliably. In outdoor settings, single‐beam laser sensors on e-scooters track rear vehicles and warn riders of collisions. 
Table \ref{tab:planning_summary} provides a visual summary for the various components, goals, proposed methods, sensors, and future directions for planning tasks in micromobility.

\section{Control}
In autonomous riding, the Control layer is responsible for executing low-level actuation commands that maintain the vehicle's upright position. This follows predetermined paths and reacts dynamically to rider inputs and environmental disruptions.  Two-wheeled platforms, in contrast to cars, are naturally unstable and need constant balancing, which makes control particularly difficult while navigating obstacles or abrupt rider motions.  This section examines current developments in balance control for e-scooters and related vehicles, covering everything from state-of-the-art deep reinforcement learning policies developed in high-fidelity simulators to traditional feedback techniques (reaction wheels, active brakes).  In order to obtain reliable, real-time stabilisation in the intricate dynamics of micromobility, we then demonstrate how these methods can be further improved by combining perceptual cues, predictive planning outputs, and multi-sensor fusion. Control is the most important part to maintain the safety of the riders. However, it has still been overlooked by the research community.  Discussion about how to progress the field of autonomous riding to catch up with the advancements of autonomous driving is provided at the end of each subsection.

\subsection{Balance}
\cite{baltes2023deep} proposed a novel deep reinforcement learning control approach that allows a humanoid robot to balance a two-wheeled scooter. It is trained entirely in an Isaac Gym \cite{nvidia2021isaacgym} using the Proximal Policy Optimization (PPO) \cite{schulman2017proximal} technique, which enables effective parallel training over thousands of scenarios. Experimental results illustrate that the DRL controller performs noticeably better than the PID controller by an average of 52\%. \cite{soloperto2021control} suggested a system that allows electric scooters to balance while riding. This helps in collision avoidance, obstacle recognition, and self-balancing. In order to enable the scooter to remain upright in the absence of a rider, especially a reaction wheel and a brake mechanism. Sensors and real-time control algorithms are used to identify impediments and modify the scooter's trajectory.  
 \cite{soloperto2021control} suggested a system that allows electric scooters to balance while riding. This helps in collision avoidance, obstacle recognition, and self-balancing. Sensors and real-time control algorithms are essential to identify impediments and modify the scooter's trajectory.

\begin{table*}[!h]
\centering
\caption{Visual summary of main functional components, techniques, sensors, and open challenges in Autonomous Micromobility balance and stabilization. This summary categorizes core capabilities including balance and dynamic stabilization. As well as, the proposed models, sensing modalities, and research opportunities that can be adapted from AD.}
\label{tab:control_summary}
\renewcommand{\arraystretch}{1.2}
\resizebox{\textwidth}{!}{%
\begin{tabular}{|l|l|l|l|l|}
\hline
\textbf{Component } & \textbf{Goal} & \textbf{Techniques/Models} & \textbf{Sensors} & \textbf{Challenges \& Future Directions} \\ \hline

\textbf{Balance Maintenance } & 
\begin{tabular}{@{}l@{}}
• Achieve static/dynamic stability \\
• Compensate for rider disturbances \\
• Maintain upright posture
\end{tabular} & 
\begin{tabular}{@{}l@{}}
• DRL-PPO (Isaac Gym) \\
• Hybrid PID-RL controllers \\
• Reaction wheel control
\end{tabular} & 
\begin{tabular}{@{}l@{}}
• 6-axis IMU (±2000°/s) \\
• Torque sensors \\
• Force-sensitive footplates
\end{tabular} & 
\begin{tabular}{@{}l@{}}
• Real-time adaptation to rider weight \\
• Predictive anti-topple algorithms \\
• Low-power edge deployment
\end{tabular} \\ \hline

\textbf{Dynamic Stabilization} & 
\begin{tabular}{@{}l@{}}
• Handle transient maneuvers \\
• Adapt to terrain changes \\
• Manage payload variations
\end{tabular} & 
\begin{tabular}{@{}l@{}}
• LQR with state estimation \\
• Neural network compensators \\
• Active brake balancing
\end{tabular} & 
\begin{tabular}{@{}l@{}}
• Millimeter-wave radar \\
• Suspension travel sensors \\
\end{tabular} & 
\begin{tabular}{@{}l@{}}
• Pothole anticipation systems \\
• Crowd-flow adaptive control \\
• Self-calibrating dynamics
\end{tabular} \\ \hline

\end{tabular}%
}
\label{tab:control}
\end{table*}
\noindent\textbf{Gaps and Directions inspirations from autonomous driving}
\\
\noindent There have been several stages in the development of autonomous driving, each characterized by advancements in sensing, control systems, and decision-making. However, balancing the vehicle while driving is a unique characteristic of micromobility as it is not present for autonomous driving. Therefore, in this section, we discuss the gaps that require the attention of the research community. Balance is a result of all the tasks of micromobility that are discussed in this survey, including path prediction, crowd counting, sensor fusion and anomaly detection. In other words, balance and safe maneuvering are not only influenced by mechanical factors but are also affected by environmental, rider interactions, and contextual variables. AI should be used to detect and react to area violations, such as uneven or restricted zones, through visual models and reinforcement learning for scene environment understanding and control, respectively. Similarly, rider-related violations, such as poor posture or distractions, can be detected using pose estimation and activity recognition models. This will allow real-time balance adjustments. Road condition analysis through surface classification helps detect potholes. Weather adversity is an important factor in predicting the rider's reactions, so that precautionary balance modes should be activated. Understanding rider activity, such as acceleration or being under the influence of drugs, through temporal models can help with proactive stabilisation. Moreover, path prediction and obstacle forecasting using deep learning can guide balance-aware trajectory planning. Finally, sensor fusion techniques that combine vision and sensing can be used as a robust perception framework. Altogether, these ML-driven modules can be integrated into reinforcement learning or hybrid control frameworks in order to equip the micromobility platform with the ability to balance in a reactive and predictive modes of the rider.
Table \ref{tab:control_summary} provides a visual summary for the various components, goals, proposed methods, sensors and future directions for control and stability tasks in micromobility.

\begin{table*}[]
\centering
\caption{Summary of Autonomous Driving  Datasets for various tasks}
\resizebox{\textwidth}{!}{
\begin{tabular}{lllll}
\hline
\textbf{Task} & \textbf{\# Datasets} & \textbf{2D/3D} & \textbf{Sensors} & \textbf{Images/Videos} \\ 
\hline
3D Object Detection & 68+ & 3D & LiDAR, cameras, radar & 1M+ images (e.g., Waymo, nuScenes) \\  
Semantic Segmentation & 52+ & Both & LiDAR, cameras & 100k+ frames (e.g., CityScapes \cite{cordts2016cityscapes}, BDD100K \cite{yu2020bdd100k}) \\  
Motion Forecasting & 25+ & 3D & LiDAR, cameras & 324k+ trajectories (e.g., Argoverse \cite{chang2019argoverse}) \\ 

Depth Estimation & 28+ & 3D & LiDAR, stereo cameras & 130k+ samples (e.g., KITTI, Gated2Depth \cite{gruber2019gated2depth}) \\  
End-to-End Driving & 12+ & Both & Cameras, GPS & 5k+ segments (e.g., Waymo E2E \cite{sun2020scalability}) \\  
 
Panoptic Segmentation & 15+ & Both & LiDAR, cameras & 78M+ points (e.g., Toronto-3D \cite{tan2020toronto}) \\  
Lane Detection & 10+ & 2D & Cameras & 50k+ images (e.g., ApolloScape \cite{huang2018apolloscape}) \\  
Multi-Object Tracking & 34+ & 3D & LiDAR, radar & 10M+ bounding boxes (e.g., nuScenes) \\  
\hline
\end{tabular}
}
\label{tab:driving_datasets}
\end{table*}

\begin{table*}[]
\centering
\caption{Summary of Micromobility Datasets for all the tasks}
\resizebox{\textwidth}{!}{
\begin{tabular}{lcccccc}
\hline
\textbf{Dataset Name / Citation} & \textbf{Task} & \textbf{Data Type} & \textbf{2D/3D} & \textbf{Size} & \textbf{Environment} & \textbf{Sensors} \\
\hline
MoRe \cite{figueiredo2021more}            & Re-identification     & Images           & 2D  & 17,619        & Outdoor            & Surveillance Cameras  \\
Gilroy et al. \cite{gilroy2022scooter}    & Detection             & Images           & 2D  & 1,130         & Outdoor            & None                  \\
Apurv et al. \cite{apurv2021detection}    & Classification        & Images           & 2D  & 21,454        & Outdoor            & None                  \\
Sabri et al. \cite{sabri2024detection}    & Detection             & Videos           & 2D  & 105           & Outdoor            & Dashcam               \\
DashCop \cite{rawat2025dashcop}           & Violation Detection   & Videos           & 2D  & 400           & Outdoor            & Dashcam, GPS          \\
LoLTV \cite{bose2023loltv}                & Anomaly Detection     & Videos           & 2D  & 105     & Nighttime Outdoor  & Dashcam               \\
iRider \cite{alwin2024irider}             & Biomechanical Analysis & Sensor + Images & 3D  & 21,454        & Outdoor            & IMU, GPS, Cameras     \\
Helmet Detection \cite{hossain2021identifying} & Helmet Detection & Images           & 2D  & 5,000         & Outdoor            & None                  \\
\hline
\end{tabular}
}
\label{tab:micromobility_datasets_2d3d}
\end{table*}

\section{Micromobility Datasets}
Despite the rapid growth of autonomous driving benchmarks supporting a lot of visual tasks, the micromobility domain remains narrowly overlooked by only a small set of vision datasets. Whereas leading autonomous-driving collections such as Waymo \cite{waymo}, nuScenes \cite{nuscenes}, and KITTI \cite{kitti} provide densely annotated 2D and 3D object detection, semantic and panoptic segmentation, multi-object tracking, motion forecasting, lane detection, depth estimation, and even end-to-end driving, micromobility has received less attention from both the research community and industry. Although a few two-wheeler datasets have been published, they are still limited to very simple tasks, including detection, re-identification, and helmet detection. Yet, no large-scale datasets exist for 3D detection, 3D segmentation, temporal tracking, or rider behaviour prediction—tasks that are more fundamental than second-order tasks such as night vision, adverse weather, or balancing rider posture. Examining Tables \ref{tab:micromobility_datasets_2d3d} and \ref{tab:driving_datasets} reveals the gaps we discuss. The minimum number of datasets dedicated to optical flow exceeds ten, significantly more than the total number of datasets available across all micromobility paradigms.

One of the main reasons is the lack of sensor diversity and data scale in micromobility collections. Autonomous-driving datasets usually fuse various types of sensors, including LiDAR, radar, stereo-vision, and GPS/IMU streams to capture rich spatial and dynamic context; however, micromobility datasets rely almost entirely on simple sensors. Moreover, while top autonomous-driving sets exceed hundreds of thousands or even millions of frames, most micromobility datasets comprise only a few thousand to tens of thousands of images or short video clips, which is insufficient to train robust deep models. Nonetheless, the absence of fine-grained annotations such as pixel-level or point-cloud-level labels and long temporal sequences is completely missing in the literature. Such gaps limited the progress of visual micromobility applications, thereby causing harm to riders' safety.

Bridging this gap will require significant attention from both the research community and industry to enable micromobility to keep pace with driving advancements. For instance, new benchmark datasets must be collected to cover the vision tasks—2D/3D detection, semantic and panoptic segmentation, multi-object tracking, and trajectory forecasting. The combination of multi-modal sensors (LiDAR, stereo or event cameras, radar, GPS/IMU) is essential to capture depth, motion, and localisation cues that are critical for perceiving riders in cluttered, dynamic scenes. Additionally, large-scale video datasets should be collected to reflect temporal tasks and anomaly detection under various adverse weather conditions. Finally, synthetic data generation via simulators like CARLA \cite{dosovitskiy2017carla} and AirSim \cite{shah2018airsim} provides a scalable approach to creating richly annotated two-wheeler scenarios. We hope these directions will establish the foundations for a robust micromobility vision task compared to autonomous-driving counterparts.

\section{Ethical and legal responsibility}
Micromobility platforms such as e-scooters and two-wheelers inherit a broad list of ethical challenges from autonomous driving. As Hansson \etal discussed in \cite{hansson2021self}, self-driving vehicles raise issues from low accident-tolerance thresholds to safety-related trade-offs, over-confidence effects, data misuse, and cybersecurity threats.  In the micromobility context, these concerns intersect with the lack of formal vehicle registration, variable rider compliance, and dense urban operation, demanding new frameworks for responsibility allocation, privacy protection, and risk mitigation \cite{swissre2021sonar}. In addition to these inherited risks of AD, various two-wheelers like e-scooters lack having license plates as they depend on app-based user IDs and GPS logs that can be combined with external datasets to re-identify individuals. This may expose detailed mobility patterns and sensitive personal behaviors that breach privacy protocols. This study \cite{vinayaga2022investigative} demonstrated that e-scooters can be remotely hacked to intercept user data, spoof GPS signals, or seize control—enabling theft, joyriding, or even misdirection into dangerous areas. Malicious actors might weaponize fleets to create public safety hazards or launch broader attacks on urban mobility networks \cite{eurekalert}.

The research community is required to address such concerns to ensure fairness and responsibility of micromobility devices. For instance, privacy-preserving analytics should be applied using various techniques, including federated learning, differential privacy, and secure multi-party computation. This will help enable real-time fleet optimization with preserving individual movement traces and sensitive rider data.  To protect two-wheelers from hacking, GPS spoofing, and remote takeover, cybersecurity methods and techniques must be taken care of, including threat modeling, penetration testing of communication protocols, and prototype intrusion-detection and tamper-resistant update architectures. Further studies are required to investigate low-cost sensor fusion to provide accurate hazard detection and adaptive speed controls.  Strategies and inclusive design adaptations, such as coverage quotas and universal audio warnings, to prevent service deserts.  Security metrics, including crash reporting, and cybersecurity incident reporting in open data platforms to enable life-cycle and behavioral analysis.

\section{Conclusion and Future Directions}

Urban transportation is changing due to the increasing presence of micromobility vehicles, such as e-bikes and e-scooters, which present complex issues in terms of road user interaction and infrastructure integration. In order to better understand how autonomous riding (AR) differs from traditional autonomous driving (AD) in important areas like perception, planning, control, and the computational constraints of lightweight platforms, this review has examined this emerging topic. Although AR can benefit from the fundamental ideas of AD, its unique characteristics, including rider fragility, vehicle instability, and hardware limitations, call for urgent attention from the research community.

Significant obstacles still exist despite promising advancements in fields such as sensor fusion, small AI models. These include installing models on devices with limited resources, frequent occlusions, and in adverse weather conditions. The absence of extensive, multi-modal datasets created especially for micromobility is one of the most urgent constraints, hindering further development in 3D visual understanding and control.

Rich sensor fusion and transformer-based models have significantly advanced AD, but AR research has not yet completely embraced these technologies. The creation of specific benchmarks that utilise various sensor modalities, including LiDAR, stereo vision, and inertial measurement units, is necessary to help address this gap. The use of reinforcement learning and physics-informed neural networks is very promising for addressing two-wheelers' balance, particularly the uniqueness of such an issue to AR rather than AD.

A persistent obstacle to the advancement of autonomous riding is the lack of large-scale, multimodal 3D datasets specifically designed for two-wheeled platforms. While autonomous driving research benefits from rich LiDAR–camera benchmarks, micromobility work remains confined to a handful of 2D image datasets. To close this gap, the community must invest in capturing synchronised LiDAR, stereo‐vision and inertial streams of e-scooters, e-bikes, and their riders across varied weather, lighting, and urban density conditions. Moreover, long temporal sequences are essential to support spatio‐temporal tasks such as trajectory forecasting and occlusion recovery, while low‐cost depth cameras and IMUs should be incorporated to reflect real‐world AR hardware constraints.

Finally, AR is poised to deliver last-mile solutions that are safe, effective, and revolutionary by taking inspiration from the development of autonomous driving and other overlapping technologies. In this survey, for the first time to the best of our knowledge, we summarised the state-of-the-art of AR over the last decade, keeping an eye on AD that will shape the future of AR. We summarised almost 60 papers in various sections, including anomaly detection, rider detection, and rider behaviour understanding. Then, a comparison between AD and AR is discussed at the end of each section, outlining the feasibility of bridging such gaps in AR. A visual comparison between datasets in AR and AD is provided to demonstrate the necessity of large-scale datasets for the safe future of micromobility.

\section*{Acknowledgement}
This research was funded by the National Road Safety Program under the Australian Government’s Department of Infrastructure, Transport, Regional Development, Communications and the Arts (Grant No. NRSAGP-TI1-A48). The authors gratefully acknowledge this funding support.

\bibliographystyle{IEEEtran}
\bibliography{main}

\begin{thebibliography}{100}
\providecommand{\url}[1]{#1}
\csname url@samestyle\endcsname
\providecommand{\newblock}{\relax}
\providecommand{\bibinfo}[2]{#2}
\providecommand{\BIBentrySTDinterwordspacing}{\spaceskip=0pt\relax}
\providecommand{\BIBentryALTinterwordstretchfactor}{4}
\providecommand{\BIBentryALTinterwordspacing}{\spaceskip=\fontdimen2\font plus
\BIBentryALTinterwordstretchfactor\fontdimen3\font minus \fontdimen4\font\relax}
\providecommand{\BIBforeignlanguage}[2]{{%
\expandafter\ifx\csname l@#1\endcsname\relax
\typeout{** WARNING: IEEEtran.bst: No hyphenation pattern has been}%
\typeout{** loaded for the language `#1'. Using the pattern for}%
\typeout{** the default language instead.}%
\else
\language=\csname l@#1\endcsname
\fi
#2}}
\providecommand{\BIBdecl}{\relax}
\BIBdecl

\bibitem{imove_micromobility}
{iMOVE Australia}, ``Micromobility,'' \url{https://imoveaustralia.com/topics/micromobility/}, 2024, accessed: 2025-05-05.

\bibitem{anke2023micro}
J.~Anke, M.~Ringhand, T.~Petzoldt, and T.~Gehlert, ``Micro-mobility and road safety: Why do e-scooter riders use the sidewalk? evidence from a german field study,'' \emph{European Transport Research Review}, vol.~15, no.~1, p.~29, 2023.

\bibitem{yaqoob2019autonomous}
I.~Yaqoob, L.~U. Khan, S.~A. Kazmi, M.~Imran, N.~Guizani, and C.~S. Hong, ``Autonomous driving cars in smart cities: Recent advances, requirements, and challenges,'' \emph{IEEE Network}, vol.~34, no.~1, pp. 174--181, 2019.

\bibitem{sanchez2024shared}
N.~C. Sanchez and K.~Larson, ``Shared autonomous micro-mobility for walkable cities,'' \emph{Transportation Research Interdisciplinary Perspectives}, vol.~27, p. 101236, 2024.

\bibitem{yu2020building}
B.~Yu, W.~Hu, L.~Xu, J.~Tang, S.~Liu, and Y.~Zhu, ``Building the computing system for autonomous micromobility vehicles: Design constraints and architectural optimizations,'' in \emph{2020 53rd Annual IEEE/ACM International Symposium on Microarchitecture (MICRO)}.\hskip 1em plus 0.5em minus 0.4em\relax IEEE, 2020, pp. 1067--1081.

\bibitem{savino2020active}
G.~Savino, R.~Lot, M.~Massaro, M.~Rizzi, I.~Symeonidis, S.~Will, and J.~Brown, ``Active safety systems for powered two-wheelers: A systematic review,'' \emph{Traffic injury prevention}, vol.~21, no.~1, pp. 78--86, 2020.

\bibitem{gail2009anti}
J.~Gail, J.~Funke, P.~Seiniger, and U.~Westerkamp, ``Anti lock braking and vehicle stability control for motorcycles-why or why not,'' in \emph{21st International Conference on the Enhanced Safety of Vehicles (ESV), Stuttgart, Germany}, 2009.

\bibitem{marra2021future}
M.~Marra, C.~Lucci, P.~Huertas-Leyva, N.~Baldanzini, M.~Pierini, and G.~Savino, ``The future of the autonomous emergency braking for powered-two-wheelers: field testing end-users’ acceptability in realistic riding manoeuvres,'' in \emph{IOP Conference Series: Materials Science and Engineering}, vol. 1038, no.~1.\hskip 1em plus 0.5em minus 0.4em\relax IOP Publishing, 2021, p. 012016.

\bibitem{sarkar1998traction}
N.~Sarkar and X.~Yun, ``Traction control of wheeled vehicles using dynamic feedback approach,'' in \emph{Proceedings. 1998 IEEE/RSJ International Conference on Intelligent Robots and Systems. Innovations in Theory, Practice and Applications (Cat. No. 98CH36190)}, vol.~1.\hskip 1em plus 0.5em minus 0.4em\relax IEEE, 1998, pp. 413--418.

\bibitem{deepan2021dynamic}
V.~Deepan, P.~Jeyakumar, and S.~Sreenath, ``Dynamic supporting wheels for two-wheeler stability,'' in \emph{Advances in Design and Thermal Systems: Select Proceedings of ETDMMT 2020}.\hskip 1em plus 0.5em minus 0.4em\relax Springer, 2021, pp. 67--79.

\bibitem{hassanin2024visual}
M.~Hassanin, S.~Anwar, I.~Radwan, F.~S. Khan, and A.~Mian, ``Visual attention methods in deep learning: An in-depth survey,'' \emph{Information Fusion}, vol. 108, p. 102417, 2024.

\bibitem{xiang2023multi}
C.~Xiang, C.~Feng, X.~Xie, B.~Shi, H.~Lu, Y.~Lv, M.~Yang, and Z.~Niu, ``Multi-sensor fusion and cooperative perception for autonomous driving: A review,'' \emph{IEEE Intelligent Transportation Systems Magazine}, vol.~15, no.~5, pp. 36--58, 2023.

\bibitem{paden2016survey}
B.~Paden, M.~{\v{C}}{\'a}p, S.~Z. Yong, D.~Yershov, and E.~Frazzoli, ``A survey of motion planning and control techniques for self-driving urban vehicles,'' \emph{IEEE Transactions on intelligent vehicles}, vol.~1, no.~1, pp. 33--55, 2016.

\bibitem{sallab2017deep}
A.~E. Sallab, M.~Abdou, E.~Perot, and S.~Yogamani, ``Deep reinforcement learning framework for autonomous driving,'' \emph{arXiv preprint arXiv:1704.02532}, 2017.

\bibitem{janai2020computer}
J.~Janai, F.~G{\"u}ney, A.~Behl, and A.~Geiger, ``Computer vision for autonomous vehicles: Problems, datasets and state-of-the-art,'' \emph{Foundations and Trends in Computer Graphics and Vision}, vol.~12, no. 1-3, pp. 1--308, 2020.

\bibitem{kaundanya2024using}
C.~Kaundanya, P.~Cesar, B.~Cronin, A.~Fleury, M.~Liu, and S.~Little, ``Using attention mechanisms in compact cnn models for improved micromobility safety through lane recognition.'' in \emph{VEHITS}, 2024, pp. 88--98.

\bibitem{livox_mid40}
{Livox Technology}, ``Livox mid‑40 lidar sensor,'' \url{https://livoxtech.com/mid-40-and-mid-100}, 2025, 38.4° circular FOV, 100 k pts/s, 260 m range.

\bibitem{titterton2004strapdown}
D.~Titterton, J.~Weston, and J.~L. Weston, \emph{Strapdown Inertial Navigation Technology}, 2nd~ed.\hskip 1em plus 0.5em minus 0.4em\relax Stevenage, UK: Institution of Engineering and Technology, 2004.

\bibitem{zheng2025real}
Z.~Zheng, A.~Hosseini, D.~Chen, O.~Shoghli, and A.~Heydarian, ``Real-time roadway obstacle detection for electric scooters using deep learning and multi-sensor fusion,'' \emph{arXiv preprint arXiv:2504.03171}, 2025.

\bibitem{prabu2022wearable}
A.~Prabu, D.~Shen, R.~Tian, S.~Chien, L.~Li, Y.~Chen, and R.~Sherony, ``A wearable data collection system for studying micro-level e-scooter behavior in naturalistic road environment,'' \emph{arXiv preprint arXiv:2212.11979}, 2022.

\bibitem{liu2018enabling}
K.~Liu and R.~Mulky, ``Enabling autonomous navigation for affordable scooters,'' \emph{Sensors}, vol.~18, no.~6, p. 1829, 2018.

\bibitem{raheel2023motorbike}
A.~Raheel, A.~Arsalan, S.~H. Noorani, S.~Khan, M.~Ehatisham-Ul-Haq, and Z.~Ali, ``Motorbike driving activity recognition using smartphone motion sensors,'' in \emph{2023 25th International Multitopic Conference (INMIC)}.\hskip 1em plus 0.5em minus 0.4em\relax IEEE, 2023, pp. 1--6.

\bibitem{alwin2024irider}
J.~Alwin, S.~J. Callista, K.~Sharon, P.~M. Kallarackal, R.~Sanjai, J.~M. Asensio-Gil, and C.~R.-M. Garc{\'\i}a, ``irider: Integrating sensors and cameras for in-depth biomechanical analysis of electric scooter,'' in \emph{2024 IEEE Applied Sensing Conference (APSCON)}.\hskip 1em plus 0.5em minus 0.4em\relax IEEE, 2024, pp. 1--4.

\bibitem{poojari2024outdoor}
S.~S. Poojari, J.~Lee, and D.~A. Paley, ``Outdoor localization and path planning for repositioning an autonomous electric scooter,'' \emph{IEEE Transactions on Intelligent Vehicles}, 2024.

\bibitem{alai2025smart}
H.~Alai, W.~Jeon, L.~Alexander, and R.~Rajamani, ``A smart e-scooter with embedded estimation of rear vehicle trajectories for rider protection,'' \emph{Mechanical Systems and Signal Processing}, vol. 222, p. 111786, 2025.

\bibitem{Luo_2024_CVPR}
X.~Luo, Z.~Cui, and F.~Su, ``Fe-det: An effective traffic object detection framework for fish-eye cameras,'' in \emph{Proceedings of the IEEE/CVF Conference on Computer Vision and Pattern Recognition (CVPR) Workshops}, June 2024, pp. 7091--7099.

\bibitem{alai2023rear}
H.~Alai, W.~Jeon, L.~Alexander, and R.~Rajamani, ``Rear vehicle tracking on a smart e-scooter,'' in \emph{2023 American Control Conference (ACC)}.\hskip 1em plus 0.5em minus 0.4em\relax IEEE, 2023, pp. 1735--1740.

\bibitem{geiger2012kitti}
A.~Geiger, P.~Lenz, and R.~Urtasun, ``Are we ready for autonomous driving? the kitti vision benchmark suite,'' in \emph{2012 IEEE Conference on Computer Vision and Pattern Recognition}.\hskip 1em plus 0.5em minus 0.4em\relax IEEE, 2012, pp. 3354--3361.

\bibitem{caesar2020nuscenes}
H.~Caesar, V.~Bankiti, A.~H. Lang, S.~Vora, V.~E. Liong, Q.~Xu, A.~Krishnan, Y.~Pan, G.~Baldan, and O.~Beijbom, ``nuscenes: A multimodal dataset for autonomous driving,'' in \emph{Proceedings of the IEEE/CVF Conference on Computer Vision and Pattern Recognition (CVPR)}, 2020, pp. 11\,621--11\,631.

\bibitem{nayak2023advances}
A.~K. Nayak, B.~Ganguli, and P.~M. Ajayan, ``Advances in electric two-wheeler technologies,'' \emph{Energy Reports}, vol.~9, pp. 3508--3530, 2023.

\bibitem{howard2017mobilenets}
A.~G. Howard, ``Mobilenets: Efficient convolutional neural networks for mobile vision applications,'' \emph{arXiv preprint arXiv:1704.04861}, 2017.

\bibitem{apurv2021detection}
K.~Apurv, R.~Tian, and R.~Sherony, ``Detection of e-scooter riders in naturalistic scenes,'' \emph{arXiv preprint arXiv:2111.14060}, 2021.

\bibitem{sandler2018mobilenetv}
M.~Sandler, A.~Howard, M.~Zhu, A.~Zhmoginov, and L.-C. Chen, ``Mobilenetv2: Inverted residuals and linear bottlenecks,'' in \emph{Proceedings of the IEEE conference on computer vision and pattern recognition}, 2018, pp. 4510--4520.

\bibitem{redmon2018yolov3}
J.~Redmon, ``Yolov3: an incremental improvement,'' in \emph{Computer Vision and Pattern Recognition}, 2018, p.~1.

\bibitem{lee2025yolov5}
S.-H. Lee, S.-H. Oh, and J.-G. Kim, ``Yolov5-based electric scooter crackdown platform,'' \emph{Applied Sciences}, vol.~15, no.~6, p. 3112, 2025.

\bibitem{zhang2022electric}
C.~Zhang, A.~Xiong, X.~Luo, C.~Zhou, and J.~Liang, ``Electric bicycle detection based on improved yolov5,'' in \emph{2022 4th International Conference on Advances in Computer Technology, Information Science and Communications (CTISC)}.\hskip 1em plus 0.5em minus 0.4em\relax IEEE, 2022, pp. 1--5.

\bibitem{gilroy2022scooter}
S.~Gilroy, D.~Mullins, E.~Jones, A.~Parsi, and M.~Glavin, ``E-scooter rider detection and classification in dense urban environments,'' \emph{Results in engineering}, vol.~16, p. 100677, 2022.

\bibitem{sabri2024detection}
K.~Sabri, C.~Djilali, G.-A. Bilodeau, N.~Saunier, and W.~Bouachir, ``Detection of micromobility vehicles in urban traffic videos,'' \emph{arXiv preprint arXiv:2402.18503}, 2024.

\bibitem{kim2024development}
C.~Kim, S.~Yu, and K.~Y. Lee, ``Development of an electric scooter photo recognition system using yolo.'' \emph{Journal of Information Processing Systems}, vol.~20, no.~6, 2024.

\bibitem{redmon2016you}
J.~Redmon, S.~Divvala, R.~Girshick, and A.~Farhadi, ``You only look once: Unified, real-time object detection,'' in \emph{Proceedings of the IEEE conference on computer vision and pattern recognition}, 2016, pp. 779--788.

\bibitem{chen2024performance}
D.~Chen, A.~Hosseini, A.~Smith, A.~F. Nikkhah, A.~Heydarian, O.~Shoghli, and B.~Campbell, ``Performance evaluation of real-time object detection for electric scooters,'' \emph{arXiv preprint arXiv:2405.03039}, 2024.

\bibitem{zuraimi2021vehicle}
M.~A.~B. Zuraimi and F.~H.~K. Zaman, ``Vehicle detection and tracking using yolo and deepsort,'' in \emph{2021 IEEE 11th IEEE Symposium on Computer Applications \& Industrial Electronics (ISCAIE)}.\hskip 1em plus 0.5em minus 0.4em\relax IEEE, 2021, pp. 23--29.

\bibitem{karagounis2024leveraging}
A.~Karagounis, ``Leveraging large language models for enhancing autonomous vehicle perception,'' \emph{arXiv preprint arXiv:2412.20230}, 2024.

\bibitem{anaya2015vulnerable}
J.~J. Anaya, E.~Talavera, D.~Gim{\'e}nez, N.~G{\'o}mez, F.~Jim{\'e}nez, and J.~E. Naranjo, ``Vulnerable road users detection using v2x communications,'' in \emph{2015 IEEE 18th international conference on intelligent transportation systems}.\hskip 1em plus 0.5em minus 0.4em\relax IEEE, 2015, pp. 107--112.

\bibitem{wang2020vulnerability}
Y.~Wang, D.~W.~M. Chia, and Y.~Ha, ``Vulnerability of deep learning model based anomaly detection in vehicle network,'' in \emph{2020 IEEE 63rd International Midwest Symposium on Circuits and Systems (MWSCAS)}.\hskip 1em plus 0.5em minus 0.4em\relax IEEE, 2020, pp. 293--296.

\bibitem{zhang2008multilevel}
W.~Zhang, Q.~J. Wu, X.~Yang, and X.~Fang, ``Multilevel framework to detect and handle vehicle occlusion,'' \emph{IEEE Transactions on Intelligent Transportation Systems}, vol.~9, no.~1, pp. 161--174, 2008.

\bibitem{coluccia2021drone}
A.~Coluccia, A.~Fascista, A.~Schumann, L.~Sommer, A.~Dimou, D.~Zarpalas, F.~C. Akyon, O.~Eryuksel, K.~A. Ozfuttu, S.~O. Altinuc \emph{et~al.}, ``Drone-vs-bird detection challenge at ieee avss2021,'' in \emph{2021 17th IEEE International Conference on Advanced Video and Signal Based Surveillance (AVSS)}.\hskip 1em plus 0.5em minus 0.4em\relax IEEE, 2021, pp. 1--8.

\bibitem{figueiredo2021more}
A.~Figueiredo, J.~Brayan, R.~O. Reis, R.~Prates, and W.~R. Schwartz, ``More: A large-scale motorcycle re-identification dataset,'' in \emph{Proceedings of the IEEE/CVF Winter Conference on Applications of Computer Vision}, 2021, pp. 4034--4043.

\bibitem{huynh2021motorbike}
C.~K. Huynh, T.~K. Dang, and C.~A. Nguyen, ``Motorbike counting in heavily crowded scenes,'' in \emph{Future Data and Security Engineering: 8th International Conference, FDSE 2021, Virtual Event, November 24--26, 2021, Proceedings 8}.\hskip 1em plus 0.5em minus 0.4em\relax Springer, 2021, pp. 175--194.

\bibitem{zheng2016person}
L.~Zheng, Y.~Yang, and A.~G. Hauptmann, ``Person re-identification: Past, present and future,'' \emph{arXiv preprint arXiv:1610.02984}, 2016.

\bibitem{wang2025vehicle}
Y.~Wang, R.~Li, and Y.~Shao, ``Vehicle re-identification method based on efficient self-attention cnn-transformer and multi-task learning optimization,'' \emph{Sensors}, vol.~25, no.~10, p. 2977, 2025.

\bibitem{khorramshahi2023robust}
P.~Khorramshahi, V.~Shenoy, and R.~Chellappa, ``Robust and scalable vehicle re-identification via self-supervision,'' in \emph{Proceedings of the IEEE/CVF Conference on Computer Vision and Pattern Recognition}, 2023, pp. 5295--5304.

\bibitem{li2024day}
H.~Li, J.~Chen, A.~Zheng, Y.~Wu, and Y.~Luo, ``Day-night cross-domain vehicle re-identification,'' in \emph{Proceedings of the IEEE/CVF Conference on Computer Vision and Pattern Recognition}, 2024, pp. 12\,626--12\,635.

\bibitem{hu2024tanet}
W.~Hu, H.~Zhan, P.~Shivakumara, U.~Pal, and Y.~Lu, ``Tanet: Text region attention learning for vehicle re-identification,'' \emph{Engineering Applications of Artificial Intelligence}, vol. 133, p. 108448, 2024.

\bibitem{wang2024vision}
D.~Wang, Q.~Wang, Z.~Tu, W.~Min, X.~Xiong, Y.~Zhong, and D.~Gai, ``Vision-language constraint graph representation learning for unsupervised vehicle re-identification,'' \emph{Expert Systems with Applications}, vol. 255, p. 124495, 2024.

\bibitem{xu2025clip}
J.~Xu, Q.~Wang, X.~Xiong, D.~Gai, R.~Zhou, and D.~Wang, ``Clip-driven view-aware prompt learning for unsupervised vehicle re-identification,'' in \emph{Proceedings of the AAAI Conference on Artificial Intelligence}, vol.~39, no.~8, 2025, pp. 8896--8904.

\bibitem{albouchi2024implementation}
A.~Albouchi, S.~Messaoud, S.~Bouaafia, M.~A. Hajjaji, and A.~Mtibaa, ``Implementation of an improved multi-object detection, tracking, and counting for autonomous driving,'' \emph{Multimedia Tools and Applications}, vol.~83, no.~18, pp. 53\,467--53\,495, 2024.

\bibitem{dai2019video}
Z.~Dai, H.~Song, X.~Wang, Y.~Fang, X.~Yun, Z.~Zhang, and H.~Li, ``Video-based vehicle counting framework,'' \emph{IEEE access}, vol.~7, pp. 64\,460--64\,470, 2019.

\bibitem{guerrero2015extremely}
R.~Guerrero-G{\'o}mez-Olmedo, B.~Torre-Jim{\'e}nez, R.~L{\'o}pez-Sastre, S.~Maldonado-Basc{\'o}n, and D.~Onoro-Rubio, ``Extremely overlapping vehicle counting,'' in \emph{Pattern Recognition and Image Analysis: 7th Iberian Conference, IbPRIA 2015, Santiago de Compostela, Spain, June 17-19, 2015, Proceedings 7}.\hskip 1em plus 0.5em minus 0.4em\relax Springer, 2015, pp. 423--431.

\bibitem{xiang2018vehicle}
X.~Xiang, M.~Zhai, N.~Lv, and A.~El~Saddik, ``Vehicle counting based on vehicle detection and tracking from aerial videos,'' \emph{Sensors}, vol.~18, no.~8, p. 2560, 2018.

\bibitem{taghvaeeyan2013portable}
S.~Taghvaeeyan and R.~Rajamani, ``Portable roadside sensors for vehicle counting, classification, and speed measurement,'' \emph{IEEE Transactions on Intelligent Transportation Systems}, vol.~15, no.~1, pp. 73--83, 2013.

\bibitem{radford2021learning}
A.~Radford, J.~W. Kim, C.~Hallacy, A.~Ramesh, G.~Goh, S.~Agarwal, G.~Sastry, A.~Askell, P.~Mishkin, J.~Clark \emph{et~al.}, ``Learning transferable visual models from natural language supervision,'' in \emph{International conference on machine learning}.\hskip 1em plus 0.5em minus 0.4em\relax PmLR, 2021, pp. 8748--8763.

\bibitem{caron2021emerging}
M.~Caron, H.~Touvron, I.~Misra, H.~J{\'e}gou, J.~Mairal, P.~Bojanowski, and A.~Joulin, ``Emerging properties in self-supervised vision transformers,'' in \emph{Proceedings of the IEEE/CVF international conference on computer vision}, 2021, pp. 9650--9660.

\bibitem{liu2016ssd}
W.~Liu, D.~Anguelov, D.~Erhan, C.~Szegedy, S.~Reed, C.-Y. Fu, and A.~C. Berg, ``Ssd: Single shot multibox detector,'' in \emph{Computer Vision--ECCV 2016: 14th European Conference, Amsterdam, The Netherlands, October 11--14, 2016, Proceedings, Part I 14}.\hskip 1em plus 0.5em minus 0.4em\relax Springer, 2016, pp. 21--37.

\bibitem{girshick2015fast}
R.~Girshick, ``Fast r-cnn,'' in \emph{Proceedings of the IEEE international conference on computer vision}, 2015, pp. 1440--1448.

\bibitem{hossain2021identifying}
M.~I. Hossain, R.~B. Muhib, and A.~Chakrabarty, ``Identifying bikers without helmets using deep learning models,'' in \emph{2021 Digital Image Computing: Techniques and Applications (DICTA)}.\hskip 1em plus 0.5em minus 0.4em\relax IEEE, 2021, pp. 01--08.

\bibitem{smith2007overview}
R.~Smith, ``An overview of the tesseract ocr engine,'' in \emph{Ninth international conference on document analysis and recognition (ICDAR 2007)}, vol.~2.\hskip 1em plus 0.5em minus 0.4em\relax IEEE, 2007, pp. 629--633.

\bibitem{tran2023robust}
D.~N.-N. Tran, L.~H. Pham, H.-J. Jeon, H.-H. Nguyen, H.-M. Jeon, T.~H.-P. Tran, and J.~W. Jeon, ``Robust automatic motorcycle helmet violation detection for an intelligent transportation system,'' in \emph{Proceedings of the IEEE/CVF Conference on Computer Vision and Pattern Recognition}, 2023, pp. 5341--5349.

\bibitem{jocher2020ultralytics}
G.~Jocher, A.~Stoken, J.~Borovec, L.~Changyu, A.~Hogan, L.~Diaconu, J.~Poznanski, L.~Yu, P.~Rai, R.~Ferriday \emph{et~al.}, ``ultralytics/yolov5: v3. 0,'' \emph{Zenodo}, 2020.

\bibitem{akhtar2024real}
A.~Akhtar, R.~Ahmed, M.~H. Yousaf, and S.~A. Velastin, ``Real-time motorbike detection: Ai on the edge perspective,'' \emph{Mathematics}, vol.~12, no.~7, p. 1103, 2024.

\bibitem{duong2023helmet}
V.~H. Duong, Q.~H. Tran, H.~S.~P. Nguyen, D.~Q. Nguyen, and T.~C. Nguyen, ``Helmet rule violation detection for motorcyclists using a custom tracking framework and advanced object detection techniques,'' in \emph{Proceedings of the IEEE/CVF Conference on Computer Vision and Pattern Recognition}, 2023, pp. 5381--5390.

\bibitem{wojke2017simple}
N.~Wojke, A.~Bewley, and D.~Paulus, ``Simple online and realtime tracking with a deep association metric,'' in \emph{2017 IEEE international conference on image processing (ICIP)}.\hskip 1em plus 0.5em minus 0.4em\relax IEEE, 2017, pp. 3645--3649.

\bibitem{hernandez2024computer}
N.~Hern{\'a}ndez-D{\'\i}az, Y.~C. Pe{\~n}aloza, Y.~Y. Rios, J.~C. Martinez-Santos, and E.~Puertas, ``A computer vision system for detecting motorcycle violations in pedestrian zones,'' \emph{Multimedia Tools and Applications}, pp. 1--24, 2024.

\bibitem{rawat2025dashcop}
D.~Rawat, K.~Gupta, A.~B. Roy, and R.~K. Sarvadevabhatla, ``Dashcop: Automated e-ticket generation for two-wheeler traffic violations using dashcam videos,'' in \emph{2025 IEEE/CVF Winter Conference on Applications of Computer Vision (WACV)}.\hskip 1em plus 0.5em minus 0.4em\relax IEEE, 2025, pp. 5387--5397.

\bibitem{goyal2022detecting}
A.~Goyal, D.~Agarwal, A.~Subramanian, C.~Jawahar, R.~K. Sarvadevabhatla, and R.~Saluja, ``Detecting, tracking and counting motorcycle rider traffic violations on unconstrained roads,'' in \emph{Proceedings of the IEEE/CVF conference on computer vision and pattern recognition}, 2022, pp. 4303--4312.

\bibitem{zhang2024coarse}
H.~Zhang, Z.~Cui, and F.~Su, ``A coarse-to-fine two-stage helmet detection method for motorcyclists,'' in \emph{Proceedings of the IEEE/CVF Conference on Computer Vision and Pattern Recognition}, 2024, pp. 7066--7074.

\bibitem{wang2023internimage}
W.~Wang, J.~Dai, Z.~Chen, Z.~Huang, Z.~Li, X.~Zhu, X.~Hu, T.~Lu, L.~Lu, H.~Li \emph{et~al.}, ``Internimage: Exploring large-scale vision foundation models with deformable convolutions,'' in \emph{Proceedings of the IEEE/CVF conference on computer vision and pattern recognition}, 2023, pp. 14\,408--14\,419.

\bibitem{wang2023prb}
B.-S. Wang, P.-Y. Chen, Y.-K. Hsieh, J.-W. Hsieh, M.-C. Chang, J.~He, S.-Y. Teng, H.~Yue, and Y.-C. Tseng, ``Prb-fpn+: Video analytics for enforcing motorcycle helmet laws,'' in \emph{Proceedings of the IEEE/CVF Conference on Computer Vision and Pattern Recognition}, 2023, pp. 5477--5485.

\bibitem{van2024motorcyclist}
T.~Van~Luong, H.~S.~P. Nguyen, D.~K. Dinh, V.~H. Duong, D.~H.~S. Vo, H.~Vu, M.~T. Hoang, and T.~C. Nguyen, ``Motorcyclist helmet violation detection framework by leveraging robust ensemble and augmentation methods,'' in \emph{Proceedings of the IEEE/CVF Conference on Computer Vision and Pattern Recognition}, 2024, pp. 7027--7036.

\bibitem{zong2023detrs}
Z.~Zong, G.~Song, and Y.~Liu, ``Detrs with collaborative hybrid assignments training,'' in \emph{Proceedings of the IEEE/CVF international conference on computer vision}, 2023, pp. 6748--6758.

\bibitem{tan2020efficientdet}
M.~Tan, R.~Pang, and Q.~V. Le, ``Efficientdet: Scalable and efficient object detection,'' in \emph{Proceedings of the IEEE/CVF conference on computer vision and pattern recognition}, 2020, pp. 10\,781--10\,790.

\bibitem{kim2024helmet}
B.-i. Kim, B.~C. Ko, I.-s. Jang, and K.-J. Kim, ``Helmet detection of motobike riders in real-world scenarios,'' in \emph{2024 IEEE International Conference on Consumer Electronics-Asia (ICCE-Asia)}.\hskip 1em plus 0.5em minus 0.4em\relax IEEE, 2024, pp. 1--4.

\bibitem{solovyev2021weighted}
R.~Solovyev, W.~Wang, and T.~Gabruseva, ``Weighted boxes fusion: Ensembling boxes from different object detection models,'' \emph{Image and Vision Computing}, vol. 107, p. 104117, 2021.

\bibitem{ahn2023cuda}
S.~Ahn, J.~Ko, and S.-Y. Yun, ``Cuda: Curriculum of data augmentation for long-tailed recognition,'' \emph{arXiv preprint arXiv:2302.05499}, 2023.

\bibitem{vo2024robust}
H.~Vo, S.~Tran, D.~M. Nguyen, T.~Nguyen, T.~Do, D.-D. Le, and T.~D. Ngo, ``Robust motorcycle helmet detection in real-world scenarios: Using co-detr and minority class enhancement,'' in \emph{Proceedings of the IEEE/CVF Conference on Computer Vision and Pattern Recognition}, 2024, pp. 7163--7171.

\bibitem{mallela2021detection}
N.~C. Mallela, R.~Volety, and N.~RK, ``Detection of the triple riding and speed violation on two-wheelers using deep learning algorithms,'' \emph{Multimedia Tools and Applications}, vol.~80, no.~6, pp. 8175--8187, 2021.

\bibitem{charran2022two}
R.~S. Charran and R.~K. Dubey, ``Two-wheeler vehicle traffic violations detection and automated ticketing for indian road scenario,'' \emph{IEEE transactions on intelligent transportation systems}, vol.~23, no.~11, pp. 22\,002--22\,007, 2022.

\bibitem{srilekha2022detection}
B.~Srilekha, K.~Kiran, and V.~V.~P. Padyala, ``Detection of license plate numbers and identification of non-helmet riders using yolo v2 and ocr method,'' in \emph{2022 International Conference on Electronics and Renewable Systems (ICEARS)}.\hskip 1em plus 0.5em minus 0.4em\relax IEEE, 2022, pp. 1539--1549.

\bibitem{bose2023loltv}
S.~Bose, M.~H. Kolekar, S.~Nawale, and D.~Khut, ``Loltv: A low light two-wheeler violation dataset with anomaly detection technique,'' \emph{IEEE Access}, vol.~11, pp. 124\,951--124\,961, 2023.

\bibitem{stallkamp2012man}
J.~Stallkamp, M.~Schlipsing, J.~Salmen, and C.~Igel, ``Man vs. computer: Benchmarking machine learning algorithms for traffic sign recognition,'' \emph{Neural networks}, vol.~32, pp. 323--332, 2012.

\bibitem{n2025real}
G.~K. N~G, A.~Kishore, A.~J. Krishna \emph{et~al.}, ``Real-time traffic sign recognition and autonomous vehicle control system using convolutional neural networks,'' \emph{Multimedia Tools and Applications}, pp. 1--36, 2025.

\bibitem{youssouf2022traffic}
N.~Youssouf, ``Traffic sign classification using cnn and detection using faster-rcnn and yolov4,'' \emph{Heliyon}, vol.~8, no.~12, 2022.

\bibitem{siebert2023computer}
F.~W. Siebert, C.~Riis, K.~H. Janstrup, H.~Lin, and F.~B. H{\"u}ttel, ``Computer vision-based helmet use registration for e-scooter riders--the impact of the mandatory helmet law in copenhagen,'' \emph{Journal of safety research}, vol.~87, pp. 257--265, 2023.

\bibitem{zhu2020understanding}
R.~Zhu, X.~Zhang, D.~Kondor, P.~Santi, and C.~Ratti, ``Understanding spatio-temporal heterogeneity of bike-sharing and scooter-sharing mobility,'' \emph{Computers, Environment and Urban Systems}, vol.~81, p. 101483, 2020.

\bibitem{brunner2020analysis}
P.~Brunner, A.~L{\"o}cken, F.~Denk, R.~Kates, and W.~Huber, ``Analysis of experimental data on dynamics and behavior of e-scooter riders and applications to the impact of automated driving functions on urban road safety,'' in \emph{2020 IEEE Intelligent Vehicles Symposium (IV)}.\hskip 1em plus 0.5em minus 0.4em\relax IEEE, 2020, pp. 219--225.

\bibitem{kegalle2025watch}
H.~N. Kegalle, D.~Hettiachchi, J.~Chan, M.~Sanderson, and F.~D. Salim, ``Watch out! e-scooter coming through!: Multimodal sensing of mixed traffic use and conflicts through riders' ego-centric views,'' \emph{Proceedings of the ACM on Interactive, Mobile, Wearable and Ubiquitous Technologies}, vol.~9, no.~1, pp. 1--23, 2025.

\bibitem{nguyen2024remote}
T.-D. Nguyen, C.~Zhang, M.~Gitbumrungsin, A.~Raheja, and T.~Chen, ``Remote kinematic analysis for mobility scooter riders leveraging edge ai,'' in \emph{Proceedings of the AAAI Symposium Series}, vol.~4, no.~1, 2024, pp. 314--318.

\bibitem{tabatabaie2024beyond}
M.~Tabatabaie, S.~He, H.~Wang, and K.~G. Shin, ``Beyond" taming electric scooters": Disentangling understandings of micromobility naturalistic riding,'' \emph{Proceedings of the ACM on Interactive, Mobile, Wearable and Ubiquitous Technologies}, vol.~8, no.~3, pp. 1--24, 2024.

\bibitem{white2023factors}
E.~White, F.~Guo, S.~Han, M.~Mollenhauer, A.~Broaddus, T.~Sweeney, S.~Robinson, A.~Novotny, and R.~Buehler, ``What factors contribute to e-scooter crashes: A first look using a naturalistic riding approach,'' \emph{Journal of safety research}, vol.~85, pp. 182--191, 2023.

\bibitem{tabatabaie2023naturalistic}
M.~Tabatabaie and S.~He, ``Naturalistic e-scooter maneuver recognition with federated contrastive rider interaction learning,'' \emph{Proceedings of the ACM on Interactive, Mobile, Wearable and Ubiquitous Technologies}, vol.~6, no.~4, pp. 1--27, 2023.

\bibitem{antony2021advanced}
M.~M. Antony and R.~Whenish, ``Advanced driver assistance systems (adas),'' in \emph{Automotive Embedded Systems: Key Technologies, Innovations, and Applications}.\hskip 1em plus 0.5em minus 0.4em\relax Springer, 2021, pp. 165--181.

\bibitem{qu2024comprehensive}
F.~Qu, N.~Dang, B.~Furht, and M.~Nojoumian, ``Comprehensive study of driver behavior monitoring systems using computer vision and machine learning techniques,'' \emph{Journal of Big Data}, vol.~11, no.~1, p.~32, 2024.

\bibitem{craye2016multi}
C.~Craye, A.~Rashwan, M.~S. Kamel, and F.~Karray, ``A multi-modal driver fatigue and distraction assessment system,'' \emph{International Journal of Intelligent Transportation Systems Research}, vol.~14, pp. 173--194, 2016.

\bibitem{siddhad2025awake}
G.~Siddhad, S.~Dey, P.~P. Roy, and M.~Iwamura, ``Awake at the wheel: Enhancing automotive safety through eeg-based fatigue detection,'' in \emph{International Conference on Pattern Recognition}.\hskip 1em plus 0.5em minus 0.4em\relax Springer, 2025, pp. 340--353.

\bibitem{zheng2017multimodal}
W.-L. Zheng and B.-L. Lu, ``A multimodal approach to estimating vigilance using eeg and forehead eog,'' \emph{Journal of neural engineering}, vol.~14, no.~2, p. 026017, 2017.

\bibitem{jiao2020driver}
Y.~Jiao, Y.~Deng, Y.~Luo, and B.-L. Lu, ``Driver sleepiness detection from eeg and eog signals using gan and lstm networks,'' \emph{Neurocomputing}, vol. 408, pp. 100--111, 2020.

\bibitem{peivandi2023deep}
M.~Peivandi, S.~Z. Ardabili, S.~Sheykhivand, and S.~Danishvar, ``Deep learning for detecting multi-level driver fatigue using physiological signals: A comprehensive approach,'' \emph{Sensors}, vol.~23, no.~19, p. 8171, 2023.

\bibitem{vaswani2017attention}
A.~Vaswani, N.~Shazeer, N.~Parmar, J.~Uszkoreit, L.~Jones, A.~N. Gomez, {\L}.~Kaiser, and I.~Polosukhin, ``Attention is all you need,'' \emph{Advances in neural information processing systems}, vol.~30, 2017.

\bibitem{hassanin2022crossformer}
M.~Hassanin, A.~Khamiss, M.~Bennamoun, F.~Boussaid, and I.~Radwan, ``Crossformer: Cross spatio-temporal transformer for 3d human pose estimation,'' \emph{arXiv preprint arXiv:2203.13387}, 2022.

\bibitem{alparslan2020towards}
K.~Alparslan, Y.~Alparslan, and M.~Burlick, ``Towards evaluating driver fatigue with robust deep learning models,'' \emph{arXiv preprint arXiv:2007.08453}, 2020.

\bibitem{yu2024driver}
L.~Yu, X.~Yang, H.~Wei, J.~Liu, and B.~Li, ``Driver fatigue detection using ppg signal, facial features, head postures with an lstm model,'' \emph{Heliyon}, vol.~10, no.~21, 2024.

\bibitem{lu2021can}
J.~Lu, X.~Zheng, L.~Tang, T.~Zhang, Q.~Z. Sheng, C.~Wang, J.~Jin, S.~Yu, and W.~Zhou, ``Can steering wheel detect your driving fatigue?'' \emph{IEEE Transactions on Vehicular Technology}, vol.~70, no.~6, pp. 5537--5550, 2021.

\bibitem{li2024intention}
C.~Li, Z.~Liu, S.~Lin, Y.~Wang, and X.~Zhao, ``Intention-convolution and hybrid-attention network for vehicle trajectory prediction,'' \emph{Expert Systems with Applications}, vol. 236, p. 121412, 2024.

\bibitem{rosmann2017kinodynamic}
C.~R{\"o}smann, F.~Hoffmann, and T.~Bertram, ``Kinodynamic trajectory optimization and control for car-like robots,'' in \emph{2017 IEEE/RSJ International Conference on Intelligent Robots and Systems (IROS)}.\hskip 1em plus 0.5em minus 0.4em\relax IEEE, 2017, pp. 5681--5686.

\bibitem{veo2025website}
\BIBentryALTinterwordspacing
{Veo}, ``Veo micromobility,'' 2025, accessed: 2025-04-29. [Online]. Available: \url{https://www.veoride.com/}
\BIBentrySTDinterwordspacing

\bibitem{mulky2018autonomous}
R.~S. Mulky, S.~Koganti, S.~Shahi, and K.~Liu, ``Autonomous scooter navigation for people with mobility challenges,'' in \emph{2018 IEEE International Conference on Cognitive Computing (ICCC)}.\hskip 1em plus 0.5em minus 0.4em\relax IEEE, 2018, pp. 87--90.

\bibitem{10711284}
S.~S. Poojari, J.~Lee, and D.~A. Paley, ``Outdoor localization and path planning for repositioning an autonomous electric scooter,'' \emph{IEEE Transactions on Intelligent Vehicles}, pp. 1--9, 2024.

\bibitem{zhang2024intent}
Z.~Zhang, Z.~Ding, Y.~Chen, S.~Chien, L.~Li, R.~Sheroy, J.~Domeyer, and R.~Tian, ``Intent-guided trajectory prediction for e-scooter riders and bicyclists,'' in \emph{2024 IEEE 27th International Conference on Intelligent Transportation Systems (ITSC)}.\hskip 1em plus 0.5em minus 0.4em\relax IEEE, 2024, pp. 272--277.

\bibitem{rudenko2020human}
A.~Rudenko, L.~Palmieri, M.~Herman, K.~M. Kitani, D.~M. Gavrila, and K.~O. Arras, ``Human motion trajectory prediction: A survey,'' \emph{The International Journal of Robotics Research}, vol.~39, no.~8, pp. 895--935, 2020.

\bibitem{Lin2022Attention}
L.~Lin, W.~Li, H.~Bi, and L.~Qin, ``Vehicle trajectory prediction using {LSTM}s with spatial-temporal attention mechanisms,'' \emph{IEEE Intelligent Transportation Systems Magazine}, vol.~14, no.~2, pp. 197–--208, 2022.

\bibitem{wang2024lstm}
J.~Wang, K.~Liu, and H.~Li, ``Lstm-based graph attention network for vehicle trajectory prediction,'' \emph{Computer Networks}, vol. 248, p. 110477, 2024.

\bibitem{singh2022multi}
D.~Singh and R.~Srivastava, ``Multi-scale graph-transformer network for trajectory prediction of the autonomous vehicles,'' \emph{Intelligent Service Robotics}, vol.~15, no.~3, pp. 307--320, 2022.

\bibitem{rossi2021vehicle}
L.~Rossi, A.~Ajmar, M.~Paolanti, and R.~Pierdicca, ``Vehicle trajectory prediction and generation using lstm models and gans,'' \emph{Plos one}, vol.~16, no.~7, p. e0253868, 2021.

\bibitem{cao2021spectral}
D.~Cao, J.~Li, H.~Ma, and M.~Tomizuka, ``Spectral temporal graph neural network for trajectory prediction,'' in \emph{2021 IEEE International Conference on Robotics and Automation (ICRA)}.\hskip 1em plus 0.5em minus 0.4em\relax IEEE, 2021, pp. 1839--1845.

\bibitem{li2024trajectory}
Z.~Li and H.~Yu, ``Trajectory prediction for autonomous driving using a transformer network,'' \emph{arXiv preprint arXiv:2402.16501}, 2024.

\bibitem{geng2023physics}
M.~Geng, J.~Li, Y.~Xia, and X.~M. Chen, ``A physics-informed transformer model for vehicle trajectory prediction on highways,'' \emph{Transportation research part C: emerging technologies}, vol. 154, p. 104272, 2023.

\bibitem{baltes2023deep}
J.~Baltes, G.~Christmann, and S.~Saeedvand, ``A deep reinforcement learning algorithm to control a two-wheeled scooter with a humanoid robot,'' \emph{Engineering Applications of Artificial Intelligence}, vol. 126, p. 106941, 2023.

\bibitem{nvidia2021isaacgym}
NVIDIA, ``Introducing isaac gym: Accelerated reinforcement learning for robotics,'' \url{https://developer.nvidia.com/blog/introducing-isaac-gym-rl-for-robotics/}, June 2021, accessed: 2025-04-29.

\bibitem{schulman2017proximal}
J.~Schulman, F.~Wolski, P.~Dhariwal, A.~Radford, and O.~Klimov, ``Proximal policy optimization algorithms,'' \emph{arXiv preprint arXiv:1707.06347}, 2017.

\bibitem{soloperto2021control}
R.~Soloperto, P.~Wenzelburger, D.~Meister, D.~Scheuble, V.~S. Breidohr, and F.~Allg{\"o}wer, ``A control framework for autonomous e-scooters,'' \emph{IFAC-PapersOnLine}, vol.~54, no.~2, pp. 252--258, 2021.

\bibitem{cordts2016cityscapes}
M.~Cordts, M.~Omran, S.~Ramos, T.~Rehfeld, M.~Enzweiler, R.~Benenson, U.~Franke, S.~Roth, and B.~Schiele, ``The cityscapes dataset for semantic urban scene understanding,'' in \emph{Proceedings of the IEEE conference on computer vision and pattern recognition}, 2016, pp. 3213--3223.

\bibitem{yu2020bdd100k}
F.~Yu, H.~Chen, X.~Wang, W.~Xian, Y.~Chen, F.~Liu, V.~Madhavan, and T.~Darrell, ``Bdd100k: A diverse driving dataset for heterogeneous multitask learning,'' in \emph{Proceedings of the IEEE/CVF conference on computer vision and pattern recognition}, 2020, pp. 2636--2645.

\bibitem{chang2019argoverse}
M.-F. Chang, J.~Lambert, P.~Sangkloy, J.~Singh, S.~Bak, A.~Hartnett, D.~Wang, P.~Carr, S.~Lucey, D.~Ramanan \emph{et~al.}, ``Argoverse: 3d tracking and forecasting with rich maps,'' in \emph{Proceedings of the IEEE/CVF conference on computer vision and pattern recognition}, 2019, pp. 8748--8757.

\bibitem{gruber2019gated2depth}
T.~Gruber, F.~Julca-Aguilar, M.~Bijelic, and F.~Heide, ``Gated2depth: Real-time dense lidar from gated images,'' in \emph{Proceedings of the IEEE/CVF International Conference on Computer Vision}, 2019, pp. 1506--1516.

\bibitem{sun2020scalability}
P.~Sun, H.~Kretzschmar, X.~Dotiwalla, A.~Chouard, V.~Patnaik, P.~Tsui, J.~Guo, Y.~Zhou, Y.~Chai, B.~Caine \emph{et~al.}, ``Scalability in perception for autonomous driving: Waymo open dataset,'' in \emph{Proceedings of the IEEE/CVF conference on computer vision and pattern recognition}, 2020, pp. 2446--2454.

\bibitem{tan2020toronto}
W.~Tan, N.~Qin, L.~Ma, Y.~Li, J.~Du, G.~Cai, K.~Yang, and J.~Li, ``Toronto-3d: A large-scale mobile lidar dataset for semantic segmentation of urban roadways,'' in \emph{Proceedings of the IEEE/CVF conference on computer vision and pattern recognition workshops}, 2020, pp. 202--203.

\bibitem{huang2018apolloscape}
X.~Huang, X.~Cheng, Q.~Geng, B.~Cao, D.~Zhou, P.~Wang, Y.~Lin, and R.~Yang, ``The apolloscape dataset for autonomous driving,'' in \emph{Proceedings of the IEEE conference on computer vision and pattern recognition workshops}, 2018, pp. 954--960.

\bibitem{waymo}
P.~Sun, H.~Kretzschmar, X.~Dotiwalla, A.~Chouard, V.~Patnaik, P.~Tsui, J.~Guo, Y.~Zhou, Y.~Chai, B.~Caine \emph{et~al.}, ``Scalability in perception for autonomous driving: Waymo open dataset,'' in \emph{Proceedings of the IEEE/CVF conference on computer vision and pattern recognition}, 2020, pp. 2446--2454.

\bibitem{nuscenes}
H.~Caesar, V.~Bankiti, A.~H. Lang, S.~Vora, V.~E. Liong, Q.~Xu, A.~Krishnan, Y.~Pan, G.~Baldan, and O.~Beijbom, ``nuscenes: A multimodal dataset for autonomous driving,'' in \emph{Proceedings of the IEEE/CVF conference on computer vision and pattern recognition}, 2020, pp. 11\,621--11\,631.

\bibitem{kitti}
Q.-H. Pham, P.~Sevestre, R.~S. Pahwa, H.~Zhan, C.~H. Pang, Y.~Chen, A.~Mustafa, V.~Chandrasekhar, and J.~Lin, ``A* 3d dataset: Towards autonomous driving in challenging environments,'' in \emph{2020 IEEE International conference on Robotics and Automation (ICRA)}.\hskip 1em plus 0.5em minus 0.4em\relax IEEE, 2020, pp. 2267--2273.

\bibitem{dosovitskiy2017carla}
A.~Dosovitskiy, G.~Ros, F.~Codevilla, A.~Lopez, and V.~Koltun, ``Carla: An open urban driving simulator,'' in \emph{Conference on robot learning}.\hskip 1em plus 0.5em minus 0.4em\relax PMLR, 2017, pp. 1--16.

\bibitem{shah2018airsim}
S.~Shah, D.~Dey, C.~Lovett, and A.~Kapoor, ``Airsim: High-fidelity visual and physical simulation for autonomous vehicles,'' in \emph{Field and Service Robotics: Results of the 11th International Conference}.\hskip 1em plus 0.5em minus 0.4em\relax Springer, 2018, pp. 621--635.

\bibitem{hansson2021self}
S.~O. Hansson, M.-{\AA}. Belin, and B.~Lundgren, ``Self-driving vehicles—an ethical overview,'' \emph{Philosophy \& Technology}, vol.~34, no.~4, pp. 1383--1408, 2021.

\bibitem{swissre2021sonar}
\BIBentryALTinterwordspacing
{Swiss Re Institute}. (2021) Electric scooters: micro-mobility, macro risk? [Online]. Available: \url{https://www.swissre.com/institute/research/sonar/sonar2021/electric-scooters.html}
\BIBentrySTDinterwordspacing

\bibitem{vinayaga2022investigative}
N.~Vinayaga-Sureshkanth, R.~Wijewickrama, A.~Maiti, and M.~Jadliwala, ``An investigative study on the privacy implications of mobile e-scooter rental apps,'' in \emph{Proceedings of the 15th ACM conference on Security and Privacy in Wireless and Mobile Networks}, 2022, pp. 125--139.

\bibitem{eurekalert}
\BIBentryALTinterwordspacing
{eurekalert}. (2020) The great e-scooter hack. [Online]. Available: \url{https://www.eurekalert.org/news-releases/627418}
\BIBentrySTDinterwordspacing

\end{thebibliography}

\end{document}